\documentclass[10pt,letterpaper,oneside,twocolumn]{article}
\usepackage[left=0.875in,right=0.875in,textheight=9.25in,top=1in,headheight=10pt,headsep=15pt,footskip=18pt]{geometry}
\usepackage{microtype}
\usepackage{flafter}
\usepackage[round,authoryear]{natbib}

\usepackage{amsmath,amssymb,amsfonts,amsthm}
\newtheorem{assumption}{Assumption}

\usepackage{algorithm}
\usepackage{algorithmic}
\usepackage{graphicx}
\usepackage{textcomp}
\usepackage{url}
\usepackage{xcolor}
\usepackage{booktabs}
\usepackage{multirow}
\usepackage{bm}
\usepackage{xspace}
\usepackage{tikz}
\usetikzlibrary{positioning, fit, arrows.meta, calc, backgrounds, decorations.pathreplacing}
\pgfdeclarelayer{background}
\pgfsetlayers{background,main}

\usepackage[hidelinks]{hyperref}
\hypersetup{
  pdftitle={Proximal Causal Learning under Unmeasured Confounding},
  pdfauthor={Ying Tang and Yi Wang},
  pdfsubject={Proximal causal learning under unmeasured confounding},
  pdfkeywords={causal inference, proximal causal learning, representation learning}
}

\begin{document}

\twocolumn[
\begin{center}
{\LARGE\bfseries Proximal Causal Learning under Unmeasured Confounding\par}
\vspace{1em}
{\large Ying Tang \quad Yi Wang\textsuperscript{*}\par}
\vspace{0.5em}
Shanghai University of International Business and Economics, Shanghai 201620, China\\
\texttt{24350012@suibe.edu.cn wangyi@suibe.edu.cn}
\vspace{1.5em}
\end{center}
]
\begingroup
\renewcommand{\thefootnote}{\fnsymbol{footnote}}
\footnotetext[1]{Corresponding author: Yi Wang (\texttt{wangyi@suibe.edu.cn}).}
\endgroup

\begin{abstract}
Estimating treatment effects from observational data typically relies on the No Unmeasured Confounding Assumption (NUCA), which rarely holds in practice. Proximal causal learning (PCL) addresses unmeasured confounding via proxy variables, yet existing methods require the proxy variables to be pre-specified. Thus, we propose PCL-U, a framework that learns proxy variables directly from observed covariates. PCL-U uses neural encoders to decompose covariates into treatment-inducing, outcome-inducing, and shared proxies, guided by minimax mutual information objectives, and obtains causal estimates through a practical moment-based risk function. Experiments on benchmarks show that PCL-U matches or outperforms existing baselines. Besides, there are two types of synthetic datasets with varying dimensions and confounding strengths that illustrate that our method maintains stable estimation accuracy.
\end{abstract}

\section{Introduction}

Causal inference is widely used to support decision-making in medicine, economics, and public policy \citep{rubin1974estimating, imbens2015causal, athey2019machine}. A central goal is to estimate treatment effects from data. While randomized controlled trials (RCTs) identify such effects by design, they are often unethical, prohibitively expensive, or infeasible \citep{bottou2013counterfactual}, motivating the use of observational data \citep{pearl2009causal}. Observational studies, however, suffer from two challenges: each unit reveals only one potential outcome \citep{imbens2015causal}, and confounders induce distributional imbalance between treated and control groups, causing selection bias \citep{rosenbaum1983central}. The problem becomes severe when some confounders are unmeasured: such latent variables affect both treatment and outcome but are absent from the data, violating NUCA and yielding biased estimates \citep{miao2018confounding}.

Recent representation-learning methods \citep{shalit2017estimating, wu2022learning, you2025disentangle} address observed confounding by learning balanced representations, but these methods require the NUCA \citep{rosenbaum1983central}. These methods cannot verify that the learned representation captures all confounding \citep{wu2022learning}, NUCA is untestable and often unrealistic in practice.

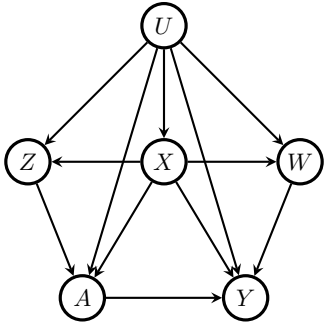
\begin{figure}[!t]
\centering
\begin{tikzpicture}[scale=0.9, transform shape, >=stealth, every node/.style={draw, circle, very thick, minimum size=6.5mm, inner sep=1pt}, thick]
\node (U) at (0, 2) {$U$};
\node (X) at (0, 0) {$X$};
\node (Z) at (-2, 0) {$Z$};
\node (W) at (2, 0) {$W$};
\node (A) at (-1.2, -2) {$A$};
\node (Y) at (1.2, -2) {$Y$};
\draw[->] (U) -- (Z);
\draw[->] (U) -- (X);
\draw[->] (U) -- (W);
\draw[->] (U) -- (A);
\draw[->] (U) -- (Y);
\draw[->] (X) -- (Z);
\draw[->] (X) -- (W);
\draw[->] (X) -- (A);
\draw[->] (X) -- (Y);
\draw[->] (Z) -- (A);
\draw[->] (W) -- (Y);
\draw[->] (A) -- (Y);
\end{tikzpicture}
\caption{Proximal causal graph.}
\label{fig:proximal_dag}
\end{figure}

When the NUCA is violated, \emph{Proximal causal learning} (PCL) \citep{miao2018confounding, tchetgen2020introduction} offers an alternative by exploiting proxy variables---treatment-inducing proxies $Z$, outcome-inducing proxies 
$W$, and shared covariates $X$---that carry information about the unmeasured confounder 
$U$ (Figure~\ref{fig:proximal_dag}). However, existing PCL methods assume valid proxies are known a priori \citep{xu2021deep,mastouri2021kernel,kompa2022deep,shao2025double} or prespecified using domain knowledge.

We therefore propose \textbf{P}roximal \textbf{C}ausal \textbf{L}earning under \textbf{U}nmeasured confounding (\textbf{PCL-U}), which automatically learns proxy variables from observed covariates $C$ and estimates treatment effects under unmeasured confounding. PCL-U maps covariates $C$ through three encoders to learn proxy variables $\hat{Z}$, $\hat{X}$, $\hat{W}$, guided by surrogate objectives that encourage the designated proximal roles, followed by single-stage neural estimation of potential outcomes using the NMMR-V formulation \citep{kompa2022deep}. Our contributions are:
\begin{itemize}
\item To our knowledge, this is the first work to integrate proxy learning into representation learning with observed covariates for treatment effect estimation.
\item On the IHDP benchmark, our method obtains the lowest PEHE and $\epsilon_{ATE}$ in both settings, reducing PEHE over the best baseline by 23.8\% within-sample and 18.6\% out-of-sample.
\item Two types of synthetic datasets are used to evaluate the limitations and robustness of our method in varying confounding scenarios, including dimension and strength. 
\end{itemize}
\section{Related Work}

Representation-learning methods reduce observed confounding in two main ways. Balancing-based methods align treated and control representations: CFR-MMD/WASS~\citep{shalit2017estimating} use distribution distances, SITE~\citep{yao2018representation} preserves local similarity, ABCEI~\citep{du2021adversarial} uses adversarial balancing, CBRE~\citep{zhou2022cycle} uses cycle consistency, DRRL~\citep{zeng2020double} uses entropy balancing, and CITE~\citep{li2022contrastive} uses contrastive learning. Decomposition-based methods instead separate covariates by their causal roles: DeR-CFR~\citep{wu2022learning} separates confounders, instruments, and adjustment variables, while DDRN~\citep{meng2025causal} learns causally disentangled representations. However, both groups rely on the No Unmeasured Confounding Assumption (NUCA) and can be biased when hidden confounders exist.

Proximal causal learning (PCL) instead uses proxy variables to address unmeasured confounding \citep{kuroki2014measurement, miao2018confounding, tchetgen2020introduction,cui2023semiparametric}. Its main computational task is to estimate bridge functions. Kernel methods use two-stage regression or moment restrictions \citep{mastouri2021kernel}; DFPV learns deep features in a two-stage model \citep{xu2021deep}; NMMR uses a single-stage neural moment objective \citep{kompa2022deep}; and DML-CMR reduces bias in two-stage conditional-moment estimation \citep{shao2025double}. These methods still require valid proxies to be given in advance.

Compared with representation-learning methods, PCL-U targets treatment effect estimation the proximal causal framework which can handle unmeasured confounding. Compared with existing proximal methods, it does not require proxy variables to be given in advance. Instead, PCL-U learns proxies directly from observed covariates and combines proxy learning with neural outcome bridge estimation in an end-to-end framework.

\section{Preliminaries}

\subsection{Problem Setup}

For each unit $i$, we observe a binary treatment $A_i\in\{0,1\}$, an outcome $Y_i$, and a pretreatment covariate vector $\mathbf{C}_i$ from a sample of $n$ units. The observations $\{(Y_i,A_i,\mathbf{C}_i)\}_{i=1}^{n}$ are assumed i.i.d.\ from a common distribution $P$. Let $Y(a)$ denote the potential outcome under treatment $a\in\{0,1\}$ \citep{rubin1974estimating}. Our target is to estimate the average treatment effect (ATE), $\tau = \mathbb{E}\{Y(1)-Y(0)\}$. Under the No Unmeasured Confounding Assumption (NUCA), all confounders are assumed observed, which is rarely guaranteed in practice. To identify $\tau$ without this assumption, we rely on treatment-inducing confounding proxies $Z$, outcome-inducing confounding proxies $W$ and shared proxies $X$, which satisfy:

\begin{assumption}[Consistency and positivity]\label{assum:consistency}
The observed outcome satisfies $Y=Y(A)$ a.s., and for each $a\in\{0,1\}$, $0<P(A=a\mid U,X)<1$ a.s.
\end{assumption}

\begin{assumption}[Latent exchangeability and conditional independence]\label{assum:exchangeability}
For each $a\in\{0,1\}$, the treatment assignment is conditionally ignorable given $(U,X)$, namely $Y(a) \perp A \mid U,X$, and the proxies satisfy $Y \perp Z \mid A,U,X$ and $W \perp (A,Z) \mid U,X$.
\end{assumption}

\begin{assumption}[Completeness]\label{assum:completeness}
For any square-integrable function $g$ and each $a\in\{0,1\}$, $\mathbb{E}\{g(U)\mid Z,A=a,X\}=0 \Rightarrow g(U)=0$ and $\mathbb{E}\{g(W)\mid Z,A=a,X\}=0 \Rightarrow g(W)=0$almost surely.
\end{assumption}

Assumption~1 states the causal consistency condition and the overlap requirement, ensuring that each unit has a positive probability of receiving either treatment level \citep{imbens2015causal}. Assumption~2 replaces the standard No Unmeasured Confounding Assumption with a proximal structure. The latent exchangeability condition $Y(a)\perp A\mid U,X$ states that, conditioning on both the unobserved confounder $U$ and the shared covariates $X$, treated and control units are exchangeable. The exclusion restriction $Y\perp Z\mid A,U,X$ requires that $Z$ affects the outcome only through its influence on treatment, while $W\perp(A,Z)\mid U,X$ requires that $W$ is excluded from the treatment mechanism given $(U,X)$. Assumption~3 imposes completeness conditions that accommodate
both categorical and continuous confounders. We note that the completeness rules out conditional independence of $U$ and $Z$ given $(A,X)$. Completeness requires sufficient proxy information about $U$. For our continuous synthetic settings,
$\min(d_z,d_w)\geq d_u$ is treated as a structural heuristic, rather than a universal implication of completeness \citep{cui2023semiparametric}.

\subsection{PCL Identification and Estimation}

Assuming that a square-integrable outcome bridge function
$h$ satisfying Eq.~\eqref{eq:outcome_bridge_section2} exists,
Assumptions~1--3 identify the counterfactual mean
$\mathbb{E}[Y(a)]$. The second completeness condition in
Assumption~3 ensures that the solution to the following
outcome bridge equation is unique up to almost-sure
equality \citep{miao2018confounding,cui2023semiparametric}:
\begin{equation}
\mathbb{E}[Y \mid Z, A, X] = \int h(w, A, X)\,dF(w \mid Z, A, X)
\label{eq:outcome_bridge_section2}
\end{equation}
The bridge functions are solved inversely through the Fredholm integral equation, which allows for the identification of the counterfactual with the presence of unmeasured confounding \citep{miao2018confounding, tchetgen2020introduction}. We defer the details of obtaining $h(W,a,X)$ to Section~4.3. For convenience, we write $h_a(W,X)\equiv h(W,a,X)$. The bridge functions provide equivalent identification formulas for the counterfactual mean:
\begin{equation}
\begin{split}
\mathbb{E}\{Y(a)\}
&= \mathbb{E}\!\left[h_a(W,X)\right]
\end{split}
\label{eq:proximal_identification_section2}
\end{equation}
The average treatment effect is identified as:
\begin{equation}
\begin{split}
\tau
&= \mathbb{E}\!\left[h_1(W,X)-h_0(W,X)\right]
\end{split}
\label{eq:ate_identification_section2}
\end{equation}
And according to \citet{cui2023semiparametric}, the outcome bridge estimator is:
\begin{equation}
\begin{split}
\psi_{\mathrm{POR}}(O;h)
&= h_1(W,X)-h_0(W,X)
\end{split}
\label{eq:proximal_dr_score}
\end{equation}
Under Assumptions~1--3, we have $\mathbb{E}\{\psi_{\mathrm{POR}}(O;h)\}=\tau$.

Existing PCL methods generally assume that valid proxy variables are known a priori. Motivated by prior work that use the learned representation to estimate causal effects\cite{wu2022learning,yao2018representation}, we introduce a novel algorithm that automatically learns proxy variables $(\hat{Z}, \hat{X}, \hat{W})$ from the observed covariates $\mathbf{C}$, fits neural bridge functions $\hat{h}$ via maximum moment restriction using these learned proxy representations. The procedure is described in Section~4.

\section{Method}\label{sec:method}

We propose \textbf{PCL-U} (Proximal
Causal Learning under Unmeasured confounding), an end-to-end framework that (i)~learns proxy representations $(\hat{Z}, \hat{X}, \hat{W})$ from the observed covariate vector $\mathbf{C}$, and (ii)~estimates bridge functions Eq.~\eqref{eq:outcome_bridge_section2} from the learned proxy representations. Throughout this section, $Z$, $X$, and $W$ denote the theoretical proximal variables in the identification analysis, whereas $\hat{Z}$, $\hat{X}$, and $\hat{W}$ denote the learned proxy representations derived from $\mathbf{C}$. We now describe the designated roles of the learned representations, motivated by the theoretical relationships in Assumption~2:

\noindent$\bullet$ \textbf{Treatment-inducing proxy $\hat{Z}$.}
The designated role of the treatment proxy is to retain information statistically associated with the treatment assignment $A$ while providing no additional information about $Y$ conditional on the treatment $A$, the unmeasured confounder $U$, and the shared proxy $\hat{X}$. 

\noindent$\bullet$ \textbf{Outcome-inducing proxy $\hat{W}$.}
The designated role of the outcome proxy is to retain information statistically associated with $Y$ while being independent of $A$ conditional on the unmeasured confounder $U$ and the shared proxy $\hat{X}$. 

\noindent$\bullet$ \textbf{Shared proxy $\hat{X}$.}
The shared proxy $\hat{X}$ captures confounding information relevant to both $A$ and $Y$.

Decomposition-based methods \citep{you2025disentangle, wu2022learning, cheng2022mimdrcfr} require independent representations. However, treatment effects can still be estimated under the proximal causal framework, even when residual undecomposed information $U$ remains among the learned proxy variables, provided that the proximal identifying assumptions hold and an outcome bridge exists. Inspired by the mutual-information-based framework of \citet{xu2022autoiv} and \citet{cheng2022mimdrcfr}, we use minimax mutual-information-based relevance and exclusion surrogates to encourage the learned representations to exhibit the designated proxy roles; details are presented in Section~4.1. Because $U$ is unobserved, these conditions cannot be verified from the mutual information objectives alone.
Establishing sufficient conditions under which the proposed
MI surrogates recover valid proximal representations remains
a direction for future theoretical work. We further design two experiments to evaluate the feasibility of our method under different dimensions and strengths of $U$ (see Section~\ref{sec:synthetic_results}).

\begin{figure*}[!t]
\centering
\includegraphics[width=0.66\textwidth]{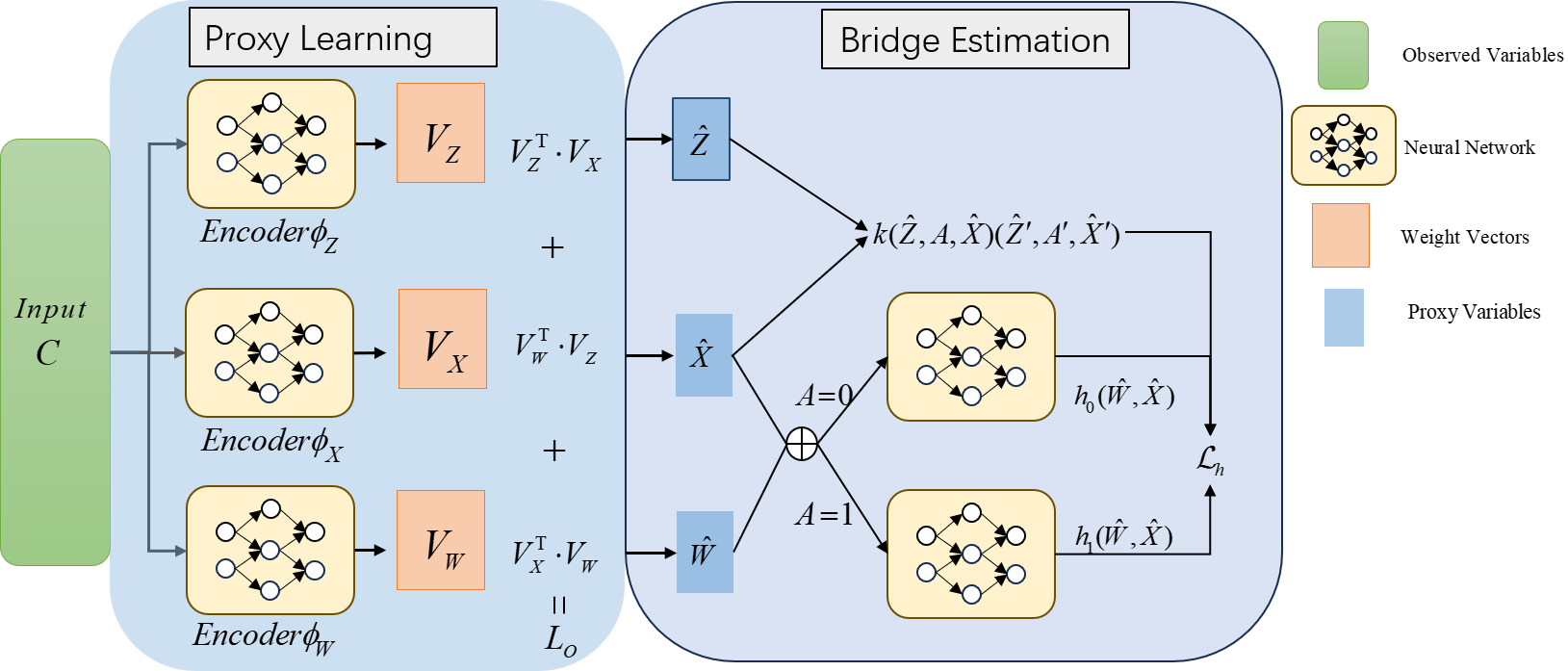}
\caption{Overview of the PCL-U architecture.}
\label{fig:framework}
\end{figure*}

\subsection{Framework}
Fig.~\ref{fig:framework} illustrates the two components of PCL-U. \textbf{Proxy learning} uses three encoder networks $\Phi_Z$, $\Phi_X$, and $\Phi_W$, trained with mutual-information-based surrogate objectives, to map the observed covariates $\mathbf{C}$ to learned proxy variables $\hat{Z}=\Phi_Z(\mathbf{C})$, $\hat{X}=\Phi_X(\mathbf{C})$, and $\hat{W}=\Phi_W(\mathbf{C})$. The vectors $V_{Z}$, $V_{X}$, and $V_{W}$ summarize the effective input weights of the encoders and are used to compute the regularization loss $L_{O}$. Thus, $\Phi_Z$, $\Phi_X$, and $\Phi_W$ denote the encoder mappings, whereas $\hat{Z}$, $\hat{X}$, and $\hat{W}$ denote their outputs used in the subsequent bridge estimation. \textbf{Bridge estimation} includes two outcome bridge networks, $( h_1(\hat{W},\hat{X})$ and $h_0(\hat{W},\hat{X}))$, and a kernel function $k(\hat{Z},A, \hat{X})(\hat{Z}',A',\hat{X}')$, which are combined to compute the bridge estimation loss $L_h$. The learned proxies are then supplied to the outcome bridges to estimate treatment effects.

\subsection{Proxy Learning}

The proxy variables $(\hat{Z}, \hat{X}, \hat{W})$ are learned from the observed covariates $C$ by combining variational conditional fitting with contrastive mutual information estimation. In this subsection, we work with the learned proxies $\hat{Z}$, $\hat{X}$, and $\hat{W}$ produced by the encoders, rather than prespecifying them using expert domain knowledge. We use a generic variable pair $\{(S, T): S\in \{\hat{Z}, \hat{W}, \hat{X}\}, T \in \{A, Y\}\}$ to show how a single loss formulation constructs observable relevance and exclusion surrogates.

We first introduce a variational conditional distribution $q_{\theta_{ST}}(T\mid S)$ parameterized by a neural network to approximate the conditional distribution $P(T\mid S)$. The log-likelihood loss of the variational approximation $q_{\theta_{ST}}(T\mid S)$ with $N$ samples is given by:
\begin{equation}
\mathcal{L}^{LLD}_{ST} = -\frac{1}{N}\sum_{i=1}^{N} \log q_{\theta_{ST}}(t_{i}\mid s_{i})
\label{eq:lld}
\end{equation}
We minimize Eq.~\eqref{eq:lld} to obtain the approximate conditional distribution $q_{\theta_{ST}}(T\mid S)$.

To construct a relevance or exclusion surrogate, we then maximize or minimize, respectively, the mutual information objective in Eq.~\eqref{eq:mi} between $S$ and $T$:
\begin{equation}
\mathcal{L}^{MI}_{ST} = \frac{1}{N^2}\sum_{i=1}^{N}\sum_{\substack{j=1\\j\neq i}}^{N}\Big(\log q_{\theta_{ST}}(t_{i}\mid s_{i}) - \log q_{\theta_{ST}}(t_{j}\mid s_{i})\Big)
\label{eq:mi}
\end{equation}
where $\log q_{\theta_{ST}}(t_i\mid s_i)$ and $\log q_{\theta_{ST}}(t_j\mid s_i)_{j\neq i}$ denote the conditional log-likelihoods of the positive pair $(s_i,t_i)$ and the negative sample pair $(s_i,t_j)_{j\neq i}$. Minimizing Eq.~\eqref{eq:mi} provides an MI-based exclusion surrogate by decreasing this difference. These surrogate objectives encourage the designated proxy roles.

We now demonstrate the procedure concretely on the treatment-inducing proxy $\hat{Z}$, and then summarize the analogous steps for $\hat{W}$ and $\hat{X}$.

\textbf{Learning $\hat{Z}$.}
The proxy-learning objective encourages $\hat{Z}$ to retain treatment-associated information while penalizing its association with $Y$.

\noindent$\bullet$ \textbf{Step 1: MI-based relevance surrogate ($\hat{Z},A$).}
We use the variational distribution $q_{\theta_{ZA}}(A\mid\hat{Z})$ with parameters $\theta_{ZA}$ to approximate the conditional distribution $P(A\mid\hat{Z})$. The log-likelihood loss is:
\begin{equation}
\mathcal{L}^{LLD}_{ZA}=-\frac{1}{N}\sum_{i=1}^{N}\log q_{\theta_{ZA}}(a_i\mid\hat{z}_i)
\label{eq:LLD_ZA}
\end{equation}
We minimize Eq.~\eqref{eq:LLD_ZA} to obtain the variational approximation $q_{\theta_{ZA}}$, and then maximize mutual information with $S=\hat{Z}$ and $T=A$:
\begin{equation}
\mathcal{L}^{MI}_{ZA}=\frac{1}{N^2}\sum_{i=1}^{N}\sum_{\substack{j=1\\j\neq i}}^{N}\!\Big(\log q_{\theta_{ZA}}(a_i\mid\hat{z}_i)-\log q_{\theta_{ZA}}(a_j\mid\hat{z}_i)\Big)
\label{eq:MI_ZA}
\end{equation}
We maximize Eq.~\eqref{eq:MI_ZA} by minimizing
$-\mathcal{L}^{MI}_{ZA}$ during gradient-based optimization, thereby providing an MI-based relevance surrogate.

\noindent$\bullet$ \textbf{Step 2: MI-based exclusion surrogate ($\hat{Z},Y$).}
We introduce the variational distribution $q_{\theta_{ZY}}(Y\mid\hat{Z})$ with parameters $\theta_{ZY}$ to approximate the conditional distribution $P(Y\mid\hat{Z})$, and the log-likelihood loss of $q_{\theta_{ZY}}$ is:
\begin{equation}
\mathcal{L}^{LLD}_{ZY}=-\frac{1}{N}\sum_{i=1}^{N}\log q_{\theta_{ZY}}(y_i\mid\hat{z}_i)
\label{eq:LLD_ZY}
\end{equation}
We minimize Eq.~\eqref{eq:LLD_ZY} to fit the variational approximation $q_{\theta_{ZY}}$. As an observable exclusion surrogate, we then minimize the mutual information objective with $S=\hat{Z}$ and $T=Y$:
\begin{equation}
\mathcal{L}^{MI}_{ZY}=\frac{1}{N^2}\sum_{i=1}^{N}\sum_{\substack{j=1\\j\neq i}}^{N}\!\Big(\log q_{\theta_{ZY}}(y_i\mid\hat{z}_i)-\log q_{\theta_{ZY}}(y_j\mid\hat{z}_i)\Big)
\label{eq:MI_ZY}
\end{equation}
Minimizing Eq.~\eqref{eq:MI_ZY} decreases the mutual information between $\hat{Z}$ and $Y$, thereby providing an MI-based exclusion surrogate.

\textbf{Learning $\hat{W}$ and $\hat{X}$.} Learning the remaining proxy variables (i.e., $\hat{W}$ and $\hat{X}$) follows the same two-step procedure: first, fitting a variational distribution to approximate the conditional distributions, and then minimizing (or maximizing) the mutual information between a proxy variable (i.e., $\hat{W}$ or $\hat{X}$) and its target (i.e., $Y$ or $A$).

Specifically, to learn the outcome-inducing proxy $\hat{W}$, we fit the variational distributions $q_{\theta_{WY}}$ and $q_{\theta_{WA}}$ to approximate the conditional distributions $P(Y\mid\hat{W})$ and $P(A\mid\hat{W})$ by minimizing $\mathcal{L}^{LLD}_{WY}$ and $\mathcal{L}^{LLD}_{WA}$, and then maximize $\mathcal{L}^{MI}_{WY}$ (equivalent to minimizing $-\mathcal{L}^{MI}_{WY}$) as an MI-based relevance surrogate with respect to $Y$ and minimize $\mathcal{L}^{MI}_{WA}$ as an MI-based exclusion surrogate with respect to $A$.

For the shared proxy $\hat{X}$, which captures confounding information relevant to both treatment $A$ and outcome $Y$, we fit the variational distributions $q_{\theta_{XA}}$ and $q_{\theta_{XY}}$ to approximate the conditional distributions $P(A\mid\hat{X})$ and $P(Y\mid\hat{X})$ by minimizing $\mathcal{L}^{LLD}_{XA}$ and $\mathcal{L}^{LLD}_{XY}$, and then maximize both $\mathcal{L}^{MI}_{XA}$ and $\mathcal{L}^{MI}_{XY}$ (equivalent to minimizing $-\mathcal{L}^{MI}_{XA}$ and $-\mathcal{L}^{MI}_{XY}$) as MI-based relevance surrogates with respect to both $A$ and $Y$.

\textbf{Regularization.} Although the proximal framework allows residual shared information across the learned proxies, the three encoders should specialize in their designated roles. Inspired by \citet{kuang2022data,wu2022learning}, we therefore regularize the parameters of the encoders $\Phi_Z$, $\Phi_X$, and $\Phi_W$. Let $R\in\{Z,X,W\}$ index an encoder, and let $\mathbf W_R^{(l)}$ denote its weight matrix at layer $l$. We define its effective input-to-layer weight matrix as $ \bar{W}_R=\mathbf W_R^{(1)}\cdots\mathbf W_R^{(L)}$ and its weight vector as $\mathbf v_R=d^{-1}|\bar{W}_R|\mathbf 1_d$, where $d$ is the width of the last constrained layer, $|\cdot|$ is applied elementwise, and $\mathbf 1_d$ is a vector of ones. Thus, each entry of $\mathbf v_R$ is the mean absolute effective weight assigned to one input covariate by encoder $\Phi_R$. We penalize overlap between the three weight vectors, and the loss is defined as:
\begin{equation}
\mathcal L_{\mathrm{O}}
=
\mathbf v_Z^\top\mathbf v_X+
\mathbf v_Z^\top\mathbf v_W+
\mathbf v_X^\top\mathbf v_W
\label{eq:orth_loss}
\end{equation}
Minimizing Eq.~\eqref{eq:orth_loss} updates the encoder parameters so that the same observed covariates are less likely to contribute strongly to multiple learned proxies, thereby promoting role-specific feature allocation. Next, we describe how to estimate counterfactual outcomes using the learned proxy variables $\hat{Z}$, $\hat{W}$, and $\hat{X}$.

\subsection{Bridge Estimation}
Deep neural networks provide flexible models for nonlinear relationships among proxies, treatment, and outcome. Following NMMR \citep{kompa2022deep}, we estimate the outcome bridge $h(W,A,X)$ by enforcing the conditional moment restriction $\mathbb{E}[Y-h(W,A,X)\mid A,X,Z]=0$. Given $\{(\hat z_i,\hat w_i,\hat x_i,a_i,y_i)\}_{i=1}^{N}$, the NMMR-V empirical loss is:
\begin{equation}
\begin{split}
\mathcal{L}_{h}={}&\frac{1}{N^2}\sum_{i,j=1}^{N}
\big\{y_i-h(\hat w_i,a_i,\hat x_i)\big\}
\big\{y_j-h(\hat w_j,a_j,\hat x_j)\big\}\\
&\quad \cdot k(\hat z_i,a_i,\hat x_i,
\hat z_j,a_j,\hat x_j)
\end{split}
\label{eq:NMMR_loss}
\end{equation}
Here, $k(G,G') = \exp\left( -\frac{\|G-G'\|_2^2}{2\ell^2} \right)$, and $G=(\hat z_i,a_i,\hat x_i)$, $G'=(\hat z_j,a_j,\hat x_j)$, and $\ell\in\{0.05,0.3\}$. We train two outcome-bridge networks, $h_0(\hat w_i,\hat x_i)$ and $h_1(\hat w_i,\hat x_i)$, and write $h(\hat w_i,a_i,\hat x_i)=(1-a_i)h_0(\hat w_i,\hat x_i)+a_ih_1(\hat w_i,\hat x_i)$ with $a_i\in\{0,1\}$. Minimizing Eq.~\eqref{eq:NMMR_loss} encourages the neural bridge networks to solve Eq.~\eqref{eq:outcome_bridge_section2}. The learned bridge gives $\hat Y_i(a)=\hat h_a(\hat W_i,\hat X_i)$, from which we estimate the average treatment effect as:
\begin{equation}
\hat{\tau}_{\mathrm{POR}} = \frac{1}{N}\sum_{i=1}^{N}\big(\hat{h}_1(\hat{W}_i,\hat{X}_i)-\hat{h}_0(\hat{W}_i,\hat{X}_i)\big)
\label{eq:ate_estimator}
\end{equation}
\subsection{Objective Function}

For clarity, we combine the loss terms. The log-likelihood losses used for approximating data distributions with variational distributions are grouped as follows:
\begin{equation}
\begin{split}
\mathcal{L}_{LLD}={}
&\mathcal{L}^{LLD}_{ZA}+\mathcal{L}^{LLD}_{ZY}+\mathcal{L}^{LLD}_{WA}\\
&+\mathcal{L}^{LLD}_{WY}+\mathcal{L}^{LLD}_{XA}+\mathcal{L}^{LLD}_{XY}
\end{split}
\label{eq:lld_loss}
\end{equation}
Accordingly, the combined mutual information loss assigns negative signs to MI-based relevance surrogate terms and positive signs to MI-based exclusion surrogate terms:
\begin{equation}
\begin{split}
\mathcal{L}_{MI}={}
&\mathcal{L}^{MI}_{ZY}-\mathcal{L}^{MI}_{ZA}+\mathcal{L}^{MI}_{WA}\\
&-\mathcal{L}^{MI}_{WY}-\mathcal{L}^{MI}_{XA}-\mathcal{L}^{MI}_{XY}
\end{split}
\label{eq:mi_loss}
\end{equation}
The proxy objective is $\mathcal{L}_{proxy}=\mathcal{L}_{LLD}+\mathcal{L}_{MI}$, and the total loss is:
\begin{equation}
\mathcal{L}_{total}=\alpha\mathcal{L}_{proxy}+\beta\mathcal{L}_{O}+\gamma\mathcal{L}_{h}+\lambda \|\theta_{h}\|^2_{2}
\label{eq:total_obj}
\end{equation}
where $\alpha$, $\beta$, $\gamma$ and $\lambda$ weight $\mathcal{L}_{\mathrm{proxy}}$, $\mathcal{L}_O$, $\mathcal{L}_h$, and the $\ell_2$ regularizer, respectively. The full Algorithm is in Appendix~\ref{sec:algorithm_appendix}.

\section{Experiment}

In this section, we demonstrate that our method remains effective for causal effect estimation in the presence of unmeasured confounding, and compare it with state-of-the-art baselines on benchmarks (IHDP, Jobs, and Twins). We further generate two families of synthetic datasets of unmeasured confounders $U$ with different dimension and strength to evaluate the limitation and effectiveness of our method. In addition, we summarize the compared baselines and datasets, introduce the evaluation metrics, and provide ablation and sensitivity analyses. Implementation details are provided in Appendix~\ref{sec:implementation_details}. The code is available.\footnote{\url{https://github.com/menchel3/PCL_U_0.git}}

\subsection{Baselines}
We compare our method \textbf{PCL-U} with nine baselines: balancing-based methods \textbf{CFR-MMD} and \textbf{CFR-WASS}~\citep{shalit2017estimating}, \textbf{DRRL}~\citep{zeng2020double}, \textbf{SITE}~\citep{yao2018representation}, \textbf{ABCEI}~\citep{du2021adversarial}, \textbf{CBRE}~\citep{zhou2022cycle}, \textbf{CITE}~\citep{li2022contrastive}, and decomposition-based methods \textbf{DeR-CFR}~\citep{wu2022learning}, \textbf{DDRN}~\citep{meng2025causal}. Additionally, we report \textbf{NMMR} \citep{kompa2022deep}, which estimates causal effects from prespecified proxy variables and uses the same way to estimate the outcome bridge function as our method.Reproduction details are provided in the accompanying code.

\subsection{Datasets}

We use \textbf{IHDP}~\citep{hill2011bayesian}, \textbf{Jobs}~\citep{lalonde1986evaluating,smith2005does}, and \textbf{Twins}~\citep{almond2005costs} because they are established benchmarks that facilitate comparison with prior treatment-effect estimation methods. Their different application domains and evaluation protocols assess the general applicability of PCL-U.

\textbf{Synthetic datasets.}
Our central claim is that PCL-U can estimate causal effects in the presence of unmeasured confounding. However, because $U$ is unobserved, the validity of proxy variables learned by neural networks cannot be assessed directly from observed data, and the presence of unmeasured confounding is itself unknown. We therefore construct synthetic datasets with known ground truth to evaluate the performance of PCL-U. The data-generating process follows the proximal causal structure in Assumption~\ref{assum:exchangeability}: $Y \perp Z \mid A,U,X$ and $W \perp (A,Z) \mid U,X$. Inspired by~\citet{zeng2020double}, we design the unmeasured confounder $U$ with varying dimensions and strengths. 

\suppressfloats[t]
\begin{table}[!t]
\centering
\footnotesize
\setlength{\tabcolsep}{2.25pt}
\renewcommand{\arraystretch}{1.05}
\caption{Summary of the datasets.}
\label{tab:dataset_summary}
\begin{tabular}{@{}ccccc@{}}
\toprule
\textbf{Property} & \textbf{IHDP} & \textbf{Jobs} & \textbf{Twins} & \textbf{Synthetic} \\
\midrule
F           & $\checkmark$ & $\checkmark$ & $\checkmark$ & $\checkmark$ \\
CF          & $\checkmark$ & $\times$     & $\checkmark$ & $\checkmark$ \\
Treatment   & Binary       & Binary       & Binary       & Binary \\
Number        & 747          & 3,212        & 11,400       & 1,000 \\
Covariates  & 25           & 17           & 30           & 12 \\
Train/valid/test & 63/27/10 & 56/24/20 & 56/24/20 & 56/24/20 \\
Replicates  & 100          & 10           & 10           & 10 \\
\bottomrule
\end{tabular}
\end{table}

\begin{table*}[!t]
\centering
\footnotesize
\setlength{\tabcolsep}{4pt}
\renewcommand{\arraystretch}{1.05}
\caption{Out-of-sample results on IHDP, Jobs and Twins.}
\label{tab:results_three_datasets}
\begin{tabular}{@{}ccccccc@{}}
\toprule
Datasets & \multicolumn{2}{c}{\textbf{IHDP}} & \multicolumn{2}{c}{\textbf{Jobs}} & \multicolumn{2}{c}{\textbf{Twins}} \\
\cmidrule(lr){2-3} \cmidrule(lr){4-5} \cmidrule(lr){6-7}
Methods & \textit{PEHE} & $\epsilon_{ATE}$ & $\mathcal{R}_{pol}$ & $\epsilon_{ATT}$ & \textit{AUC} & $\epsilon_{ATE}$ \\
\midrule
CFR-MMD  & $1.202 \pm 0.521$ & $0.275 \pm 0.120$ & $0.245 \pm 0.015$ & $\mathbf{0.064 \pm 0.026}$ & $0.540 \pm 0.013$ & $0.013 \pm 0.002$ \\
CFR-WASS & $0.804 \pm 0.257$ & $0.169 \pm 0.065$ & $0.252 \pm 0.015$ & $0.087 \pm 0.040$ & $0.511 \pm 0.015$ & $0.014 \pm 0.003$ \\
DRRL     & $2.154 \pm 0.881$ & $0.500 \pm 0.152$ & $0.256 \pm 0.015$ & $0.109 \pm 0.043$ & $0.856 \pm 0.004$ & $0.013 \pm 0.002$ \\
SITE     & $0.888 \pm 0.345$ & $0.259 \pm 0.099$ & $0.249 \pm 0.009$ & $0.067 \pm 0.030$ & $0.699 \pm 0.032$ & $0.023 \pm 0.009$ \\
ABCEI    & $1.059 \pm 0.124$ & $0.234 \pm 0.030$ & $0.345 \pm 0.006$ & $0.108 \pm 0.026$ & $0.647 \pm 0.028$ & $0.010 \pm 0.003$ \\
CBRE     & $0.616 \pm 0.063$ & $0.137 \pm 0.020$ & $0.331 \pm 0.007$ & $0.183 \pm 0.044$ & $0.686 \pm 0.049$ & $0.009 \pm 0.004$ \\
CITE     & $0.721 \pm 0.093$ & $0.135 \pm 0.021$ & $0.263 \pm 0.024$ & $0.086 \pm 0.035$ & $0.686 \pm 0.007$ & $0.016 \pm 0.002$ \\
DeR-CFR  & $0.566 \pm 0.073$ & $0.134 \pm 0.012$ & $0.249 \pm 0.013$ & $0.078 \pm 0.022$ & $0.827 \pm 0.007$ & $0.009 \pm 0.003$ \\
DDRN     & $0.813 \pm 0.158$ & $0.166 \pm 0.110$ & $\mathbf{0.211 \pm 0.020}$ & $0.069 \pm 0.022$ & $0.644 \pm 0.029$ & $0.021 \pm 0.019$ \\
\midrule
PCL-U    & $\mathbf{0.461 \pm 0.037}$ & $\mathbf{0.129 \pm 0.019}$ & $0.249 \pm 0.015$ & $0.077 \pm 0.035$ & $\mathbf{0.873 \pm 0.002}$ & $\mathbf{0.007 \pm 0.001}$ \\
\bottomrule
\end{tabular}
\end{table*}

\begin{table*}[!t]
\centering
\footnotesize
\begin{minipage}[t]{0.495\textwidth}
\vspace{0pt}
\centering
\caption{Out-of-sample results on \emph{dimension datasets} ($d_u$-$d_z$-$d_x$-$d_w$).}
\label{tab:synthetic}
\setlength{\tabcolsep}{1.5pt}
\renewcommand{\arraystretch}{1.05}
\medmuskip=0mu
\begin{tabular}{@{}cccc@{}}
\toprule
Methods & \textbf{1-4-4-4} & \textbf{2-4-4-4 (m)} & \textbf{8-4-4-4} \\
\midrule
CFR-MMD  & $0.074 \pm 0.024$ & $0.096 \pm 0.023$ & $0.332 \pm 0.072$ \\
CFR-WASS & $0.052 \pm 0.016$ & $0.096 \pm 0.023$ & $0.393 \pm 0.059$ \\
DRRL     & $0.064 \pm 0.015$ & $0.109 \pm 0.031$ & $0.422 \pm 0.066$ \\
SITE     & $0.113 \pm 0.028$ & $0.175 \pm 0.048$ & $0.479 \pm 0.064$ \\
ABCEI    & $0.057 \pm 0.018$ & $0.079 \pm 0.022$ & $0.338 \pm 0.066$ \\
CBRE     & $0.030 \pm 0.012$ & $0.045 \pm 0.019$ & $0.249 \pm 0.056$ \\
CITE     & $0.037 \pm 0.013$ & $0.071 \pm 0.023$ & $0.222 \pm 0.053$ \\
DeR-CFR  & $0.843 \pm 0.288$ & $0.527 \pm 0.164$ & $0.629 \pm 0.050$ \\
DDRN     & $0.046 \pm 0.031$ & $0.050 \pm 0.038$ & $0.275 \pm 0.204$ \\
\midrule
NMMR     & $\mathbf{0.029 \pm 0.016}$ & $0.065 \pm 0.046$ & $0.272 \pm 0.189$ \\
PCL-U    & $0.036 \pm 0.012$ & $\mathbf{0.037 \pm 0.013}$ & $\mathbf{0.210 \pm 0.057}$ \\
\bottomrule
\end{tabular}
\end{minipage}\hfill
\begin{minipage}[t]{0.495\textwidth}
\vspace{0pt}
\centering
\caption{Out-of-sample results on \emph{strength datasets}($s$, $m$, and $h$).}
\label{tab:synthetic_strength}
\setlength{\tabcolsep}{1.5pt}
\renewcommand{\arraystretch}{1.05}
\medmuskip=0mu
\begin{tabular}{@{}cccc@{}}
\toprule
Methods & \textbf{s} & \textbf{m (2-4-4-4)} & \textbf{h} \\
\midrule
CFR-MMD  & $0.033 \pm 0.012$ & $0.096 \pm 0.023$ & $0.367 \pm 0.068$ \\
CFR-WASS & $0.085 \pm 0.036$ & $0.096 \pm 0.023$ & $0.356 \pm 0.101$ \\
DRRL     & $0.030 \pm 0.007$ & $0.109 \pm 0.031$ & $0.339 \pm 0.151$ \\
SITE     & $0.078 \pm 0.029$ & $0.175 \pm 0.048$ & $0.445 \pm 0.114$ \\
ABCEI    & $0.024 \pm 0.008$ & $0.079 \pm 0.022$ & $0.314 \pm 0.114$ \\
CBRE     & $\mathbf{0.009 \pm 0.003}$ & $0.045 \pm 0.019$ & $0.195 \pm 0.056$ \\
CITE     & $0.013 \pm 0.005$ & $0.071 \pm 0.023$ & $0.231 \pm 0.087$ \\
DeR-CFR  & $0.210 \pm 0.053$ & $0.527 \pm 0.164$ & $0.766 \pm 0.216$ \\
DDRN     & $0.017 \pm 0.023$ & $0.050 \pm 0.038$ & $0.248 \pm 0.159$ \\
\midrule
NMMR     & $0.021 \pm 0.026$ & $0.065 \pm 0.046$ & $0.128 \pm 0.087$ \\
PCL-U    & $0.016 \pm 0.006$ & $\mathbf{0.037 \pm 0.013}$ & $\mathbf{0.084 \pm 0.035}$ \\
\bottomrule
\end{tabular}
\end{minipage}
\end{table*}

For the \emph{dimension datasets}: $d_u$-$d_z$-$d_x$-$d_w$. We fix
$d_z=d_x=d_w=4$ and the confounding strength $\kappa=1$, while varying
$d_u\in\{1,2,8\}$. For the \emph{strength datasets}: $s$, $m$, and $h$. We fix
$(d_u,d_z,d_x,d_w)=(2,4,4,4)$ and set
$\kappa\in\{0.5,1,2\}$ to represent small ($s$), medium ($m$), and high ($h$) levels of unmeasured
confounding, respectively. For each setting, we first sample the unmeasured confounder $U$ with
dimension $d_u$. We then generate $Z$, $X$, and $W$ as noisy functions
of $U$ and transform their concatenation using random orthogonal matrices and a nonlinear $\tanh$ perturbation to form the observed covariates $\mathbf{C}$. Treatment is generated from $U$, $Z$, and $X$, while the potential
outcomes are generated from $U$, $W$, and $X$. The parameter $\kappa$
controls the strength of the effects of $U$ on both treatment and outcome.
The factual outcome is selected according to the assigned treatment, and
the other potential outcome is retained as counterfactual ground truth.

The summary of datasets is provided in Table~\ref{tab:dataset_summary}. Further details about the three datasets and synthetic data generation are provided in Appendix~\ref{sec:dataset_details}.

\subsection{Evaluation Metrics}
We report $\epsilon_{\mathrm{PEHE}}$ and $\epsilon_{\mathrm{ATE}}$ on IHDP, policy risk and $\epsilon_{\mathrm{ATT}}$ on Jobs, AUC and $\epsilon_{\mathrm{ATE}}$ on Twins, and $\epsilon_{\mathrm{ATE}}$ on the synthetic datasets. Detailed definitions are provided in Appendix~\ref{sec:metrics_appendix}.

\subsection{Performance on Datasets}
We report the out-of-sample results discussed in this subsection as the mean
and standard deviation (std) for each evaluation metric. The corresponding
within-sample results are provided in Appendix~\ref{sec:within_sample_results}.

\subsubsection{IHDP, Jobs, and Twins}
We compare our method with nine baselines (e.g., DeR-CFR) on IHDP,
Jobs, and Twins. The results are shown in Table~\ref{tab:results_three_datasets}. Our method achieves the best results on four of the six metrics. On IHDP, it obtains the lowest PEHE and $\epsilon_{ATE}$, reducing PEHE by
18.6\% relative to the best baseline (DeR-CFR). On Jobs, DDRN and CFR-MMD achieve the best
results on $\mathcal{R}_{pol}$ and $\epsilon_{ATT}$, respectively. On Twins, PCL-U achieves the
highest AUC and the lowest $\epsilon_{ATE}$, reducing $\epsilon_{ATE}$ by 22.2\% relative to the best baseline. 

\subsubsection{Synthetic Datasets}
\label{sec:synthetic_results}
On the synthetic datasets, we report the error in average treatment effect estimation, $\epsilon_{\mathrm{ATE}}$, and vary the dimension and strength of $U$ to
evaluate our method. We compare PCL-U with NMMR to assess whether learned proxies
can achieve competitive empirical performance relative to prespecified proxies.

Table~\ref{tab:synthetic} reports the results for the \emph{dimension datasets}: $d_u$-$d_z$-$d_x$-$d_w$. We make the following observations: (1) most NUCA-based methods (e.g., CITE) achieve low errors in the 1-4-4-4 setting, while the PCL-based method NMMR achieves the best performance; (2) as the dimension of $U$ increases, our method achieves the lowest error; and (3) notably, the performance of PCL-U deteriorates in the 8-4-4-4 setting, where the latent confounder dimension exceeds each proxy dimension $d_u>\min(d_z,d_w)$ , illustrating a limitation of PCL-U.

\begin{table*}[!t]
\centering
\caption{Ablation study.}
\label{tab:ablation}
\footnotesize
\setlength{\tabcolsep}{3pt}
\renewcommand{\arraystretch}{1.05}
\begin{tabular}{@{}llccccc@{}}
\toprule
\multicolumn{2}{c}{Ablation} & \textbf{IHDP} & \textbf{1-4-4-4} & \textbf{2-4-4-4 (m)} & \textbf{s} & \textbf{h} \\
\midrule
\multirow{2}{*}{w/o $\mathcal{L}_{\mathrm{proxy}}$}
& within-sample  & $0.102 \pm 0.011$ & $0.050 \pm 0.018$ & $0.089 \pm 0.039$ & $0.055 \pm 0.015$ & $0.599 \pm 0.212$ \\
& out-of-sample & $0.136 \pm 0.021$ & $0.056 \pm 0.021$ & $0.092 \pm 0.038$ & $0.057 \pm 0.014$ & $0.599 \pm 0.213$ \\
\midrule
\multirow{2}{*}{w/o $\mathcal{L}_{O}$}
& within-sample  & $0.124 \pm 0.014$ & $\mathbf{0.032 \pm 0.014}$ & $0.045 \pm 0.014$ & $0.032 \pm 0.007$ & $0.104 \pm 0.043$ \\
& out-of-sample & $0.171 \pm 0.023$ & $0.040 \pm 0.018$ & $0.055 \pm 0.016$ & $0.034 \pm 0.007$ & $0.105 \pm 0.044$ \\
\midrule
\multirow{2}{*}{Full}
& within-sample  & $\mathbf{0.094 \pm 0.012}$ & $0.035 \pm 0.014$ & $\mathbf{0.032 \pm 0.011}$ & $\mathbf{0.021 \pm 0.008}$ & $\mathbf{0.080 \pm 0.039}$ \\
& out-of-sample & $\mathbf{0.129 \pm 0.019}$ & $\mathbf{0.036 \pm 0.012}$ & $\mathbf{0.037 \pm 0.013}$ & $\mathbf{0.016 \pm 0.006}$ & $\mathbf{0.084 \pm 0.037}$ \\
\bottomrule
\addlinespace[0.2em]
\end{tabular}
\end{table*}

\begin{figure*}[!t]
\centering
\begin{minipage}[b]{0.24\textwidth}
\centering
\includegraphics[width=\linewidth]{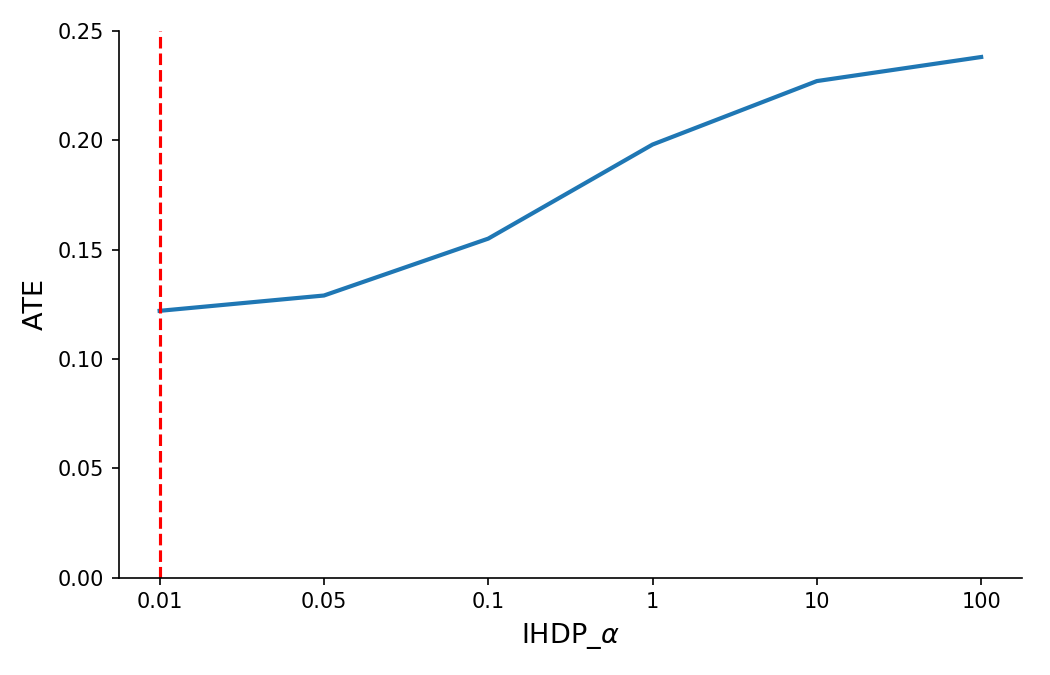}
\end{minipage}\hspace{0.05\textwidth}%
\begin{minipage}[b]{0.24\textwidth}
\centering
\includegraphics[width=\linewidth]{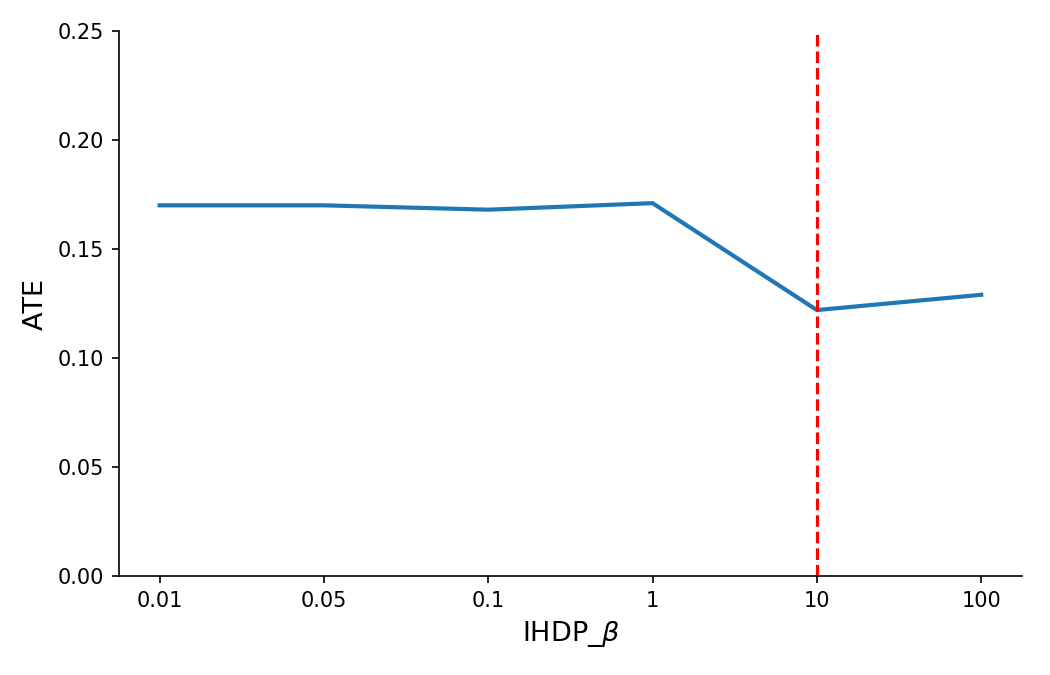}
\end{minipage}\hspace{0.05\textwidth}%
\begin{minipage}[b]{0.24\textwidth}
\centering
\includegraphics[width=\linewidth]{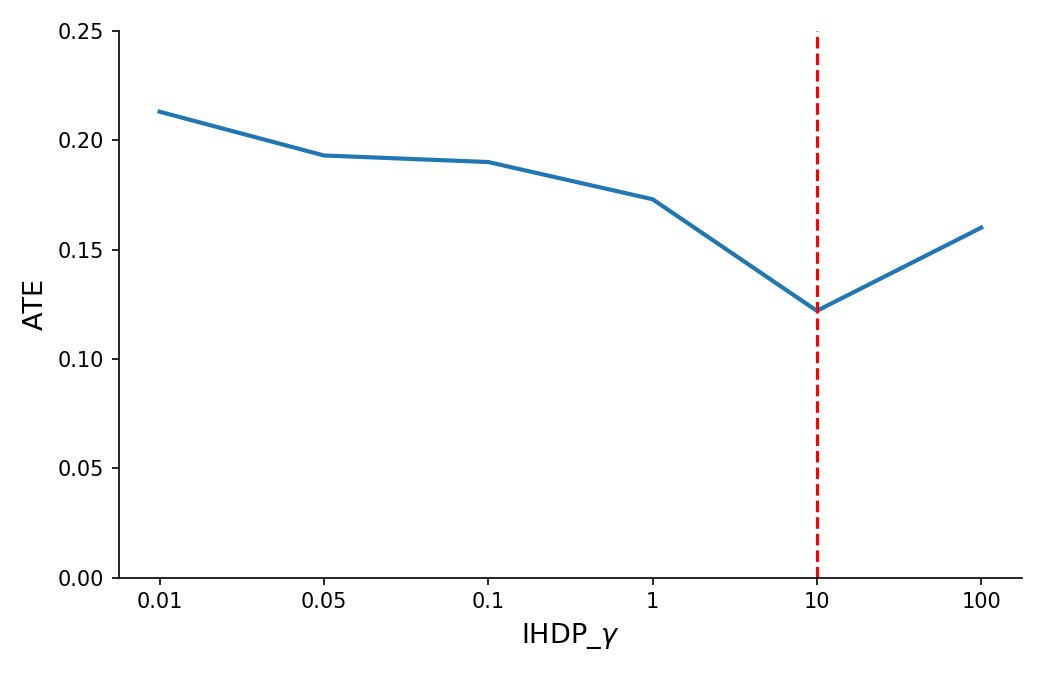}
\end{minipage}

\vspace{0.4em}
\begin{minipage}[b]{0.24\textwidth}
\centering
\includegraphics[width=\linewidth]{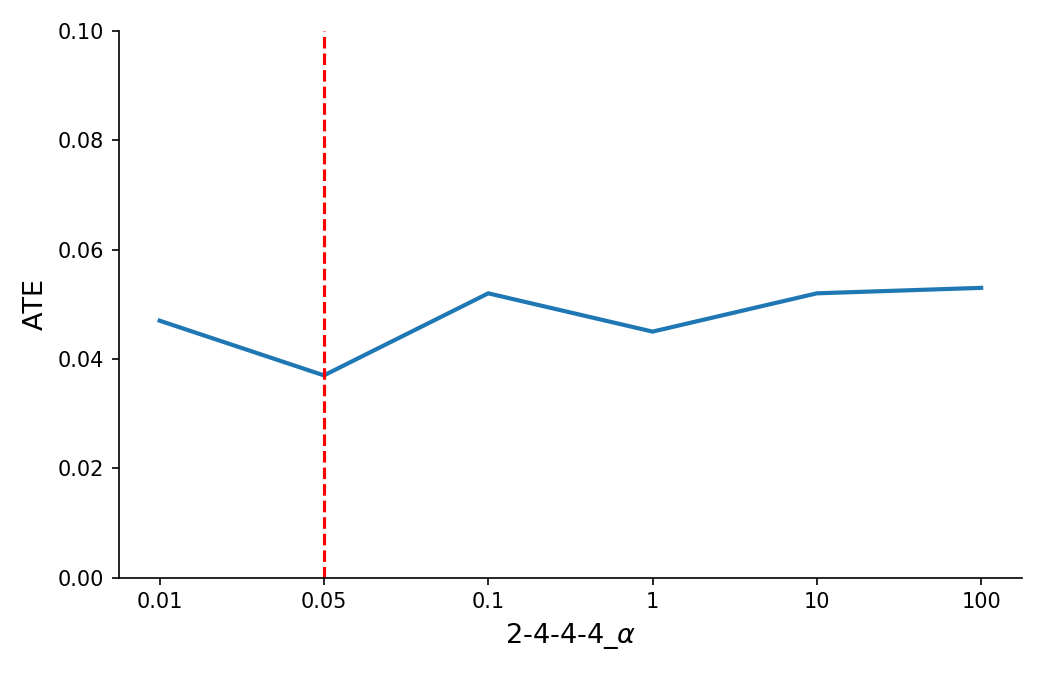}
\end{minipage}\hspace{0.05\textwidth}%
\begin{minipage}[b]{0.24\textwidth}
\centering
\includegraphics[width=\linewidth]{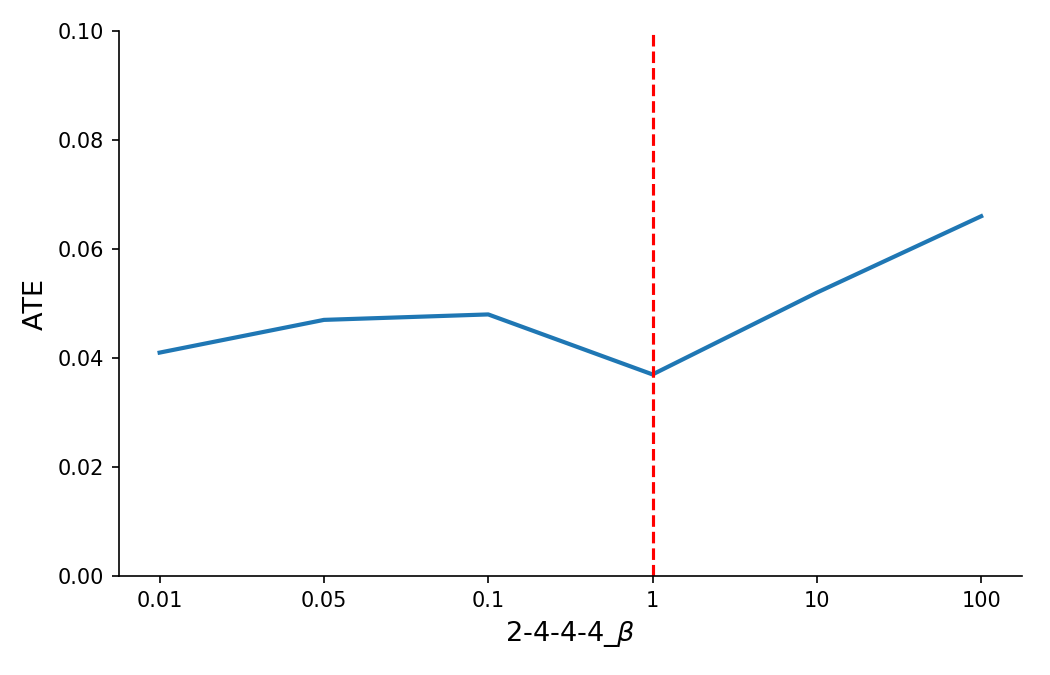}
\end{minipage}\hspace{0.05\textwidth}%
\begin{minipage}[b]{0.24\textwidth}
\centering
\includegraphics[width=\linewidth]{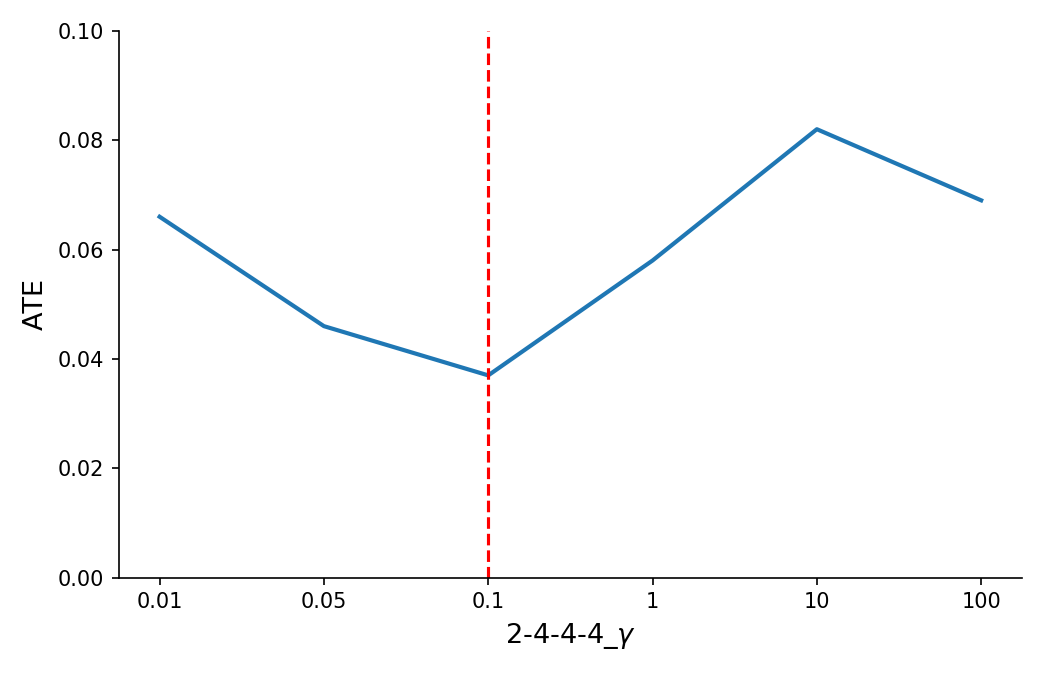}
\end{minipage}

\caption{Sensitivity of PCL-U to $\alpha$, $\beta$, and $\gamma$ on IHDP and the synthetic 2-4-4-4 setting.}
\label{fig:sensitivity}
\end{figure*}

Table~\ref{tab:synthetic_strength} reports the results for the \emph{strength datasets}: $s$, $m$, and $h$. The results show that (1) under weak confounding, CBRE achieves
the lowest error of $0.009$, while PCL-U obtains the third-lowest error of $0.016$, which is close to that of the PCL-based method NMMR; and (2) as the strength of unmeasured confounding increases, NMMR and PCL-U remain comparatively robust, and PCL-U achieves the lowest error in the $m$ and $h$ settings, reducing the error
over CBRE by 17.8\% and 56.9\%, respectively. These results show that PCL-U remains effective as the confounding strength increases.

\subsection{Ablation Study}
We examine the contribution of the combined proxy objective and orthogonality objectives to proxy learning. Table~\ref{tab:ablation} reports the
within-sample and out-of-sample $\epsilon_{\mathrm{ATE}}$ on IHDP and four synthetic settings. The results show that every component of our model is important. The only exception
is the within-sample 1-4-4-4 setting, where removing the $\mathcal{L}_{O}$ term yields a slightly lower mean error, suggesting that $\mathcal{L}_{O}$ may constrain the learning of $Z$ and $W$ when information about $U$ is limited. Overall, the combined proxy and orthogonality objectives encourage the learning of proxy variables.

\subsection{Sensitivity Analysis}
We study the sensitivity of PCL-U to $\alpha$, $\beta$, and $\gamma$, which weight $\mathcal{L}_{\mathrm{proxy}}$, $\mathcal{L}_{O}$, and $\mathcal{L}_{h}$, respectively, in Eq.~\eqref{eq:total_obj}. Specifically, we vary each hyperparameter over $\{0.01, 0.05, 0.1, 1, 10, 100\}$ while holding the others fixed. Figure~\ref{fig:sensitivity} reports the test-set $\epsilon_{\mathrm{ATE}}$ on IHDP and the synthetic 2-4-4-4 setting, while the results for the synthetic 1-4-4-4, $s$, and $h$ settings are provided in Appendix~\ref{sec:sensitivity_appendix}. As shown in Figure~\ref{fig:sensitivity}, performance changes smoothly and remains relatively stable over a broad range of values.

\section{Conclusion}
We propose PCL-U, a framework that automatically learns proxy variables from observed covariates and incorporates them into neural bridge function estimators to estimate treatment effects in the presence of the unmeasured confounder $U$. Experiments on the IHDP, Jobs, and Twins datasets show that PCL-U matches or outperforms baselines based on the No Unmeasured Confounding Assumption. We further design and evaluate synthetic datasets with varying dimensions and strengths of $U$, demonstrating the limitation imposed by the $d_u>\min(d_z,d_w)$ condition and showing that PCL-U remains effective as the confounding strength increases and achieves competitive empirical performance relative to NMMR without requiring prespecified proxies. Because the validity of learned proxies cannot be assessed directly from observed data, establishing theoretical guarantees for their identifiability remains an open problem. Future work may explore more effective and scalable methods for learning proxy variables and establishing their identifiability.

\clearpage
\bibliographystyle{apalike}
\bibliography{refs}

\clearpage
\appendix
\onecolumn
\pagestyle{plain}
\section*{Supplementary Materials}

\section{Algorithm}
\label{sec:algorithm_appendix}

\begin{algorithm}[H]
\caption{Proximal Causal Learning under Unmeasured Confounding (PCL-U)}
\label{alg:pcl_u}
\begin{algorithmic}[1]
\footnotesize
\STATE {\bf Input:} Data $\mathcal D=\{(c_i,a_i,y_i)\}_{i=1}^N$ and hyperparameters $\alpha,\beta,\gamma,\lambda$.
\STATE {\bf Initialize:} Proxy encoders $\Phi_Z,\Phi_X,\Phi_W$; variational networks $q_{\theta_{ZA}},q_{\theta_{ZY}},q_{\theta_{XA}},q_{\theta_{XY}},q_{\theta_{WA}},q_{\theta_{WY}}$; and outcome bridges $h_0,h_1$.
\FOR{$t=0,\ldots,T-1$}
    \STATE Sample a mini-batch and obtain normalized proxies $\hat Z=\Phi_Z(C)$, $\hat X=\Phi_X(C)$, and $\hat W=\Phi_W(C)$.
    \STATE Compute the proxy loss $\mathcal L_{\rm proxy}$ and regularization loss $\mathcal L_{O}$.
    \STATE Compute $\hat h=(1-A)h_0(\hat W,\hat X)+Ah_1(\hat W,\hat X)$ and the outcome bridge loss $\mathcal L_h$.
    \STATE Compute $\mathcal L_{\rm total}$ via Eq.~\eqref{eq:total_obj}.
    \STATE \parbox[t]{0.78\columnwidth}{\raggedright Jointly update encoder parameters $\theta_{\Phi_Z},\theta_{\Phi_X},\theta_{\Phi_W}$;\\variational parameters $\theta_{ZA},\theta_{ZY},\theta_{XA},\theta_{XY},\theta_{WA},\theta_{WY}$; and bridge parameters $\theta_{h_0},\theta_{h_1}$ using Adam.}
\ENDFOR
\STATE Estimate $\hat Y_i(a)=\hat h_a(\hat W_i,\hat X_i)$ and $\hat\tau_{\rm POR}$ using Eq.~\eqref{eq:ate_estimator}.
\STATE {\bf Output:} Learned proxies $\hat Z,\hat X,\hat W$, bridges $\hat h_0,\hat h_1$, and $\hat\tau_{\rm POR}$.
\end{algorithmic}
\end{algorithm}

\section{Dataset Details}
\label{sec:dataset_details}

\textbf{IHDP}~\citep{hill2011bayesian}. The IHDP dataset is a semi-synthetic benchmark based on the Infant Health and Development Program. It retains real covariates while synthesizing potential outcomes under a non-random treatment assignment. The dataset contains 747 units (139 treated and 608 control units) with 25 covariates. We employ a 63/27/10 train/validation/test split and replicate the experiment 100 times~\citep{li2022contrastive}.

\textbf{Jobs}~\citep{lalonde1986evaluating,smith2005does}. The Jobs dataset studies the effect of job training on employment status by combining a randomized experiment with an observational comparison group from the Panel Study of Income Dynamics (PSID). The randomized component from the National Supported Work program includes 722 units (425 control and 297 treated) with 17 covariates, while the PSID comparison group includes 2,490 control units, yielding 3,212 units in total. We employ a 56/24/20 train/validation/test split and replicate the experiment 10 times.

\textbf{Twins}~\citep{almond2005costs}. The Twins dataset is derived from U.S. twin births recorded between 1989 and 1991, restricted to twins weighing less than 2~kg with complete covariate records. Each sample contains 30 covariates describing the parents, pregnancy, and birth. We set treatment $A=1$ for the heavier twin and $A=0$ for the lighter twin, with one-year mortality as the outcome. The final dataset contains 11,400 twin pairs. To simulate selection bias, we follow prior work~\citep{louizos2017causal,yoon2018ganite,zhang2020learning} and retain one twin from each pair as the observed unit, with selection probability $a_i\mid x_i\sim\mathrm{Bern}(\sigma(w^{\top}x_i+n))$, where $w\sim\mathcal{U}((-0.1,0.1)^{30\times 1})$ and $n\sim\mathcal{N}(0,0.1)$. We employ a 56/24/20 train/validation/test split and replicate the experiment 10 times.

\textbf{Synthetic datasets}. The unmeasured confounder $U\in\mathbb{R}^{d_u}$ is sampled from
$U\sim\mathcal{N}(\mathbf{0},\mathbf{I}_{d_u})$, and the proxy variables are generated by
\begin{equation}
P=\frac{1}{\sqrt{d_u}}U\bm{\alpha}_P+\bm{\epsilon}_P
\qquad P\in\{Z,W,X\}.
\end{equation}
Here, $Z\in\mathbb{R}^{d_z}$, $W\in\mathbb{R}^{d_w}$,
$X\in\mathbb{R}^{d_x}$, and
$\bm{\alpha}_P\in\mathbb{R}^{d_u\times d_P}$ for
$P\in\{Z,W,X\}$.
In these data-generating equations, covariates and proxies are represented
as row vectors, while the coefficient vectors $\bm{\beta}$ and
$\bm{\delta}$ are column vectors of matching dimensions.
For $Z$ and $W$, the entries of $\bm{\alpha}_P$ are sampled from
$\mathcal{U}(-2,2)$, whereas
$\bm{\alpha}_X\sim\mathcal{U}(-0.5,0.5)$. The proxy noise vectors are
mutually independent and satisfy
$\bm{\epsilon}_P\sim\mathcal{N}(\mathbf{0},0.5^2\mathbf{I}_{d_P})$
for $P\in\{Z,W,X\}$.

To conceal the predefined proxy partition from the learning algorithm,
we construct the observed input as
\begin{equation}
\mathbf{C}
=
[Z,X,W]\mathbf{Q}_1
+0.3\tanh\!\left([Z,X,W]\mathbf{Q}_1\mathbf{Q}_2\right),
\end{equation}
where $[Z,X,W]$ concatenates the three underlying proxy variables, and
$\mathbf{Q}_1$ and $\mathbf{Q}_2$ are random orthogonal matrices
obtained by QR decomposition.

Treatment and outcome are generated according to
\begin{equation}
\begin{aligned}
A &\sim \operatorname{Bernoulli}\!\left[\sigma(\eta_A-\overline{\eta}_A)\right],\\
Y &= \mu_0(U,W,X)+A\tau(U,X)+\epsilon_Y,
\end{aligned}
\end{equation}
where $\eta_A=\frac{\kappa}{\sqrt{d_u}}U\bm{\beta}^{A}_{U}
+Z\bm{\beta}^{A}_{Z}+X\bm{\beta}^{A}_{X}$, and $\overline{\eta}_A$ denotes the sample mean of $\eta_A$. Moreover,
\begin{equation}
\begin{aligned}
\mu_0(U,W,X)
&=
\frac{\kappa}{\sqrt{d_u}}U\bm{\beta}^{Y}_{U}
+W\bm{\beta}^{Y}_{W}
+X\bm{\beta}^{Y}_{X},\\
\tau(U,X)
&=
1+
\frac{1}{\sqrt{d_u}}U\bm{\delta}_{U}
+
\frac{1}{\sqrt{d_x}}X\bm{\delta}_{X}.
\end{aligned}
\end{equation}
Here, $\epsilon_Y\sim\mathcal{N}(0,0.05^2)$,
$\bm{\beta}^{A}_{U},
\bm{\beta}^{Y}_{U}\sim\mathcal{U}(0.5,1.5)$, and $\kappa$
controls the strength of the effects of $U$ on treatment and outcome.
The remaining coefficients follow
$\bm{\beta}^{A}_{Z},\bm{\beta}^{Y}_{W}\sim\mathcal{U}(-0.5,0.5)$,
$\bm{\beta}^{A}_{X},\bm{\beta}^{Y}_{X}\sim\mathcal{U}(-0.1,0.1)$,
$\bm{\delta}_{U}\sim\mathcal{U}(0.1,0.3)$, and
$\bm{\delta}_{X}\sim\mathcal{U}(0,0.05)$.

\section{Evaluation Metrics}
\label{sec:metrics_appendix}

Since the characteristics of the datasets differ, we use different metrics across datasets. For IHDP, where counterfactual outcomes are available, we report the root precision in estimation of heterogeneous effect ($\epsilon_{\mathrm{PEHE}}$) and absolute average treatment effect error ($\epsilon_{\mathrm{ATE}}$). Specifically, $\epsilon_{\mathrm{PEHE}}=\sqrt{\frac{1}{N}\sum_{i=1}^{N}(\hat{\tau}_i-\tau_i)^2}$, where $\hat{\tau}_i=\hat{y}_i(1)-\hat{y}_i(0)$ and $\tau_i=y_i(1)-y_i(0)$. The ATE error is $\epsilon_{\mathrm{ATE}}=|\widehat{\mathrm{ATE}}-\mathrm{ATE}|$, where $\mathrm{ATE}=\frac{1}{N}\sum_{i=1}^{N}\tau_i$ and $\widehat{\mathrm{ATE}}=\frac{1}{N}\sum_{i=1}^{N}\hat{\tau}_i$. We also use $\epsilon_{\mathrm{ATE}}$ for the synthetic datasets.

For Twins, following prior work \citep{louizos2017causal, zhou2022cycle}, we report the area under the ROC curve (AUC) and $\epsilon_{\mathrm{ATE}}$.

For Jobs, where counterfactual outcomes are unavailable, we report policy risk $\mathcal{R}_{pol}=1-\widehat{V}(\pi)$ \citep{shalit2017estimating, zhang2020learning} and absolute error in the average treatment effect on the treated, $\epsilon_{\mathrm{ATT}}=|\widehat{\mathrm{ATT}}-\mathrm{ATT}|$. Here, $\pi(\mathbf{x}_i)=\mathbb{I}[\hat{\tau}_i>0]$ and $\widehat{V}(\pi)=\hat{\mu}_1(1)\hat{p}(1)+\hat{\mu}_0(0)\hat{p}(0)$, with $\hat{\mu}_a(b)=\widehat{\mathbb{E}}[y\mid A=a,\pi(\mathbf{x})=b]$ and $\hat{p}(b)=\widehat{\mathbb{P}}(\pi(\mathbf{x})=b)$. The estimated ATT is $\widehat{\mathrm{ATT}}=\frac{1}{|\mathcal{T}|}\sum_{i\in\mathcal{T}}\hat{\tau}_i$, where $\mathcal{T}$ is the treated set. Lower values indicate better performance for all metrics except AUC, for which higher values are better.

\section{Within-sample Results}
\label{sec:within_sample_results}

\subsection{IHDP, Jobs, and Twins}

\begin{table}[H]
\centering
\footnotesize
\setlength{\tabcolsep}{4pt}
\renewcommand{\arraystretch}{1.05}
\caption{Within-sample results on IHDP, Jobs and Twins.}
\label{tab:results_train_benchmark}
\begin{tabular}{@{}ccccccc@{}}
\toprule
Datasets & \multicolumn{2}{c}{\textbf{IHDP}} & \multicolumn{2}{c}{\textbf{Jobs}} & \multicolumn{2}{c}{\textbf{Twins}} \\
\cmidrule(lr){2-3} \cmidrule(lr){4-5} \cmidrule(lr){6-7}
Methods & $\textit{PEHE}$ & $\epsilon_{ATE}$ & $\mathcal{R}_{pol}$ & $\epsilon_{ATT}$ & $\textit{AUC}$ & $\epsilon_{ATE}$ \\
\midrule
CFR-MMD  & $0.692 \pm 0.090$ & $0.240 \pm 0.088$ & $0.223 \pm 0.004$ & $0.039 \pm 0.018$ & $0.549 \pm 0.010$ & $0.017 \pm 0.001$ \\
CFR-WASS & $0.508 \pm 0.046$ & $0.155 \pm 0.054$ & $0.221 \pm 0.004$ & $0.063 \pm 0.023$ & $0.516 \pm 0.013$ & $0.017 \pm 0.000$ \\
DRRL     & $1.386 \pm 0.247$ & $0.553 \pm 0.138$ & $0.233 \pm 0.003$ & $0.054 \pm 0.020$ & $0.856 \pm 0.003$ & $0.014 \pm 0.001$ \\
SITE     & $0.623 \pm 0.113$ & $0.329 \pm 0.138$ & $0.223 \pm 0.004$ & $0.045 \pm 0.018$ & $0.709 \pm 0.033$ & $0.024 \pm 0.009$ \\
ABCEI    & $0.872 \pm 0.055$ & $0.220 \pm 0.027$ & $0.217 \pm 0.001$ & $0.066 \pm 0.026$ & $0.648 \pm 0.027$ & $0.007 \pm 0.003$ \\
CBRE     & $0.503 \pm 0.032$ & $0.099 \pm 0.014$ & $0.218 \pm 0.003$ & $0.068 \pm 0.024$ & $0.693 \pm 0.043$ & $0.008 \pm 0.003$ \\
CITE     & $0.592 \pm 0.059$ & $0.096 \pm 0.013$ & $0.242 \pm 0.009$ & $0.049 \pm 0.012$ & $0.689 \pm 0.006$ & $0.015 \pm 0.002$ \\
DeR-CFR  & $0.470 \pm 0.040$ & $0.119 \pm 0.015$ & $0.180 \pm 0.005$ & $0.071 \pm 0.011$ & $0.849 \pm 0.007$ & $0.005 \pm 0.001$ \\
DDRN     & $0.705 \pm 0.150$ & $0.150 \pm 0.118$ & $\mathbf{0.121 \pm 0.017}$ & $0.065 \pm 0.039$ & $0.674 \pm 0.003$ & $0.020 \pm 0.019$ \\
\midrule
PCL-U    & $\mathbf{0.358 \pm 0.014}$ & $\mathbf{0.094 \pm 0.012}$ & $0.212 \pm 0.005$ & $\mathbf{0.029 \pm 0.012}$ & $\mathbf{0.882 \pm 0.001}$ & $\mathbf{0.002 \pm 0.001}$ \\
\bottomrule
\addlinespace[0.2em]
\end{tabular}
\end{table}

Table~\ref{tab:results_train_benchmark} summarizes the within-sample results
on IHDP, Jobs, and Twins. PCL-U ranks first on five of the six evaluation
metrics. On IHDP, it achieves a PEHE of $0.358$ and an $\epsilon_{\mathrm{ATE}}$ of
$0.094$; its PEHE is 23.8\% lower than the best baseline value of $0.470$.
For Jobs, DDRN obtains the lowest policy risk of $0.121$, whereas PCL-U gives
the lowest $\epsilon_{\mathrm{ATT}}$ of $0.029$, improving over CFR-MMD by 25.6\%.
For Twins, PCL-U achieves both the highest AUC of $0.882$ and the lowest
$\epsilon_{\mathrm{ATE}}$ of $0.002$, with the latter representing a 60.0\% reduction
relative to the best baseline.

\subsection{Synthetic Datasets}

\begin{table}[!ht]
\centering
\caption{Within-sample results on \emph{dimension datasets} ($d_u$-$d_z$-$d_x$-$d_w$).}
\label{tab:synthetic_train}
\small
\setlength{\tabcolsep}{8pt}
\renewcommand{\arraystretch}{1.05}
\begin{tabular}{@{}cccc@{}}
\toprule
Methods & \textbf{1-4-4-4} & \textbf{2-4-4-4 (m)} & \textbf{8-4-4-4} \\
\midrule
CFR-MMD  & $0.074 \pm 0.024$ & $0.097 \pm 0.023$ & $0.344 \pm 0.074$ \\
CFR-WASS & $0.046 \pm 0.013$ & $0.097 \pm 0.023$ & $0.394 \pm 0.061$ \\
DRRL     & $0.066 \pm 0.014$ & $0.107 \pm 0.031$ & $0.434 \pm 0.072$ \\
SITE     & $0.119 \pm 0.029$ & $0.172 \pm 0.050$ & $0.510 \pm 0.062$ \\
ABCEI    & $0.060 \pm 0.018$ & $0.072 \pm 0.021$ & $0.345 \pm 0.067$ \\
CBRE     & $0.029 \pm 0.011$ & $0.046 \pm 0.019$ & $0.257 \pm 0.055$ \\
CITE     & $0.037 \pm 0.012$ & $0.069 \pm 0.022$ & $0.222 \pm 0.057$ \\
DeR-CFR  & $0.809 \pm 0.280$ & $0.536 \pm 0.167$ & $0.650 \pm 0.055$ \\
DDRN     & $0.045 \pm 0.037$ & $0.052 \pm 0.033$ & $0.286 \pm 0.204$ \\
\midrule
NMMR     & $\mathbf{0.023 \pm 0.015}$ & $0.047 \pm 0.023$ & $0.263 \pm 0.177$ \\
PCL-U    & $0.035 \pm 0.014$ & $\mathbf{0.032 \pm 0.011}$ & $\mathbf{0.218 \pm 0.057}$ \\
\bottomrule
\end{tabular}
\end{table}

Table~\ref{tab:synthetic_train} reports how within-sample estimation error
changes with the dimension of $U$. In the 1-4-4-4 setting, NMMR obtains the
lowest error of $0.023$, compared with $0.035$ for PCL-U. When $d_u$ increases
to 2, PCL-U achieves the best result of $0.032$, reducing the error relative
to CBRE by 30.4\%. PCL-U remains the best-performing method in the 8-4-4-4
setting and narrowly outperforms CITE by 1.8\%. Nevertheless, its error rises
by 581.3\% from the 2-4-4-4 setting to the 8-4-4-4 setting. This deterioration illustrates the challenges of estimation
when the latent confounder dimension exceeds each proxy dimension.

\begin{table}[!ht]
\centering
\caption{Within-sample results on \emph{strength datasets}($s$, $m$, and $h$).}
\label{tab:synthetic_strength_train}
\small
\setlength{\tabcolsep}{8pt}
\renewcommand{\arraystretch}{1.05}
\begin{tabular}{@{}cccc@{}}
\toprule
Methods & \textbf{s} & \textbf{m (2-4-4-4)} & \textbf{h} \\
\midrule
CFR-MMD  & $0.031 \pm 0.012$ & $0.097 \pm 0.023$ & $0.362 \pm 0.068$ \\
CFR-WASS & $0.086 \pm 0.033$ & $0.097 \pm 0.023$ & $0.086 \pm 0.033$ \\
DRRL     & $0.031 \pm 0.007$ & $0.107 \pm 0.031$ & $0.338 \pm 0.055$ \\
SITE     & $0.067 \pm 0.025$ & $0.172 \pm 0.050$ & $0.444 \pm 0.116$ \\
ABCEI    & $0.021 \pm 0.008$ & $0.072 \pm 0.021$ & $0.299 \pm 0.109$ \\
CBRE     & $\mathbf{0.008 \pm 0.003}$ & $0.046 \pm 0.019$ & $0.194 \pm 0.058$ \\
CITE     & $0.014 \pm 0.005$ & $0.069 \pm 0.022$ & $0.225 \pm 0.083$ \\
DeR-CFR  & $0.218 \pm 0.049$ & $0.536 \pm 0.167$ & $0.770 \pm 0.218$ \\
DDRN     & $0.014 \pm 0.022$  & $0.052 \pm 0.033$ & $0.260 \pm 0.167$ \\
\midrule
NMMR     & $0.013 \pm 0.009$ & $0.047 \pm 0.023$ & $0.121 \pm 0.068$ \\
PCL-U    & $0.021 \pm 0.008$ & $\mathbf{0.032 \pm 0.011}$ & $\mathbf{0.080 \pm 0.035}$ \\
\bottomrule
\end{tabular}
\end{table}

Table~\ref{tab:synthetic_strength_train} examines within-sample performance
under different levels of unmeasured confounding. Under weak confounding,
CBRE achieves the lowest error of $0.008$, while PCL-U obtains $0.021$. Under
medium confounding, PCL-U ranks first with an error of $0.032$, which is
30.4\% lower than that of CBRE. Under strong confounding, PCL-U again achieves
the lowest error, obtaining $0.080$ and improving on the closest baseline,
CFR-WASS, by 7.0\%. Thus, PCL-U becomes comparatively more effective as the
strength of unmeasured confounding increases.

\section{Sensitivity Results}
\label{sec:sensitivity_appendix}

\begin{figure}[!ht]
\centering
\begin{minipage}[b]{0.24\textwidth}
\centering
\includegraphics[width=\linewidth]{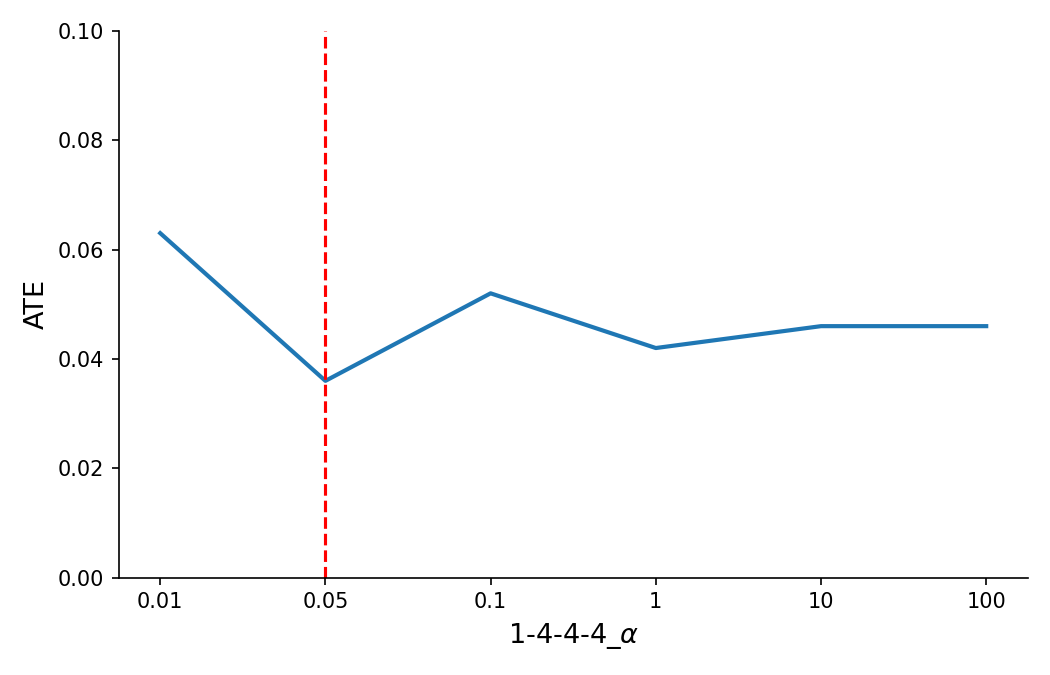}
\end{minipage}\hspace{0.05\textwidth}%
\begin{minipage}[b]{0.24\textwidth}
\centering
\includegraphics[width=\linewidth]{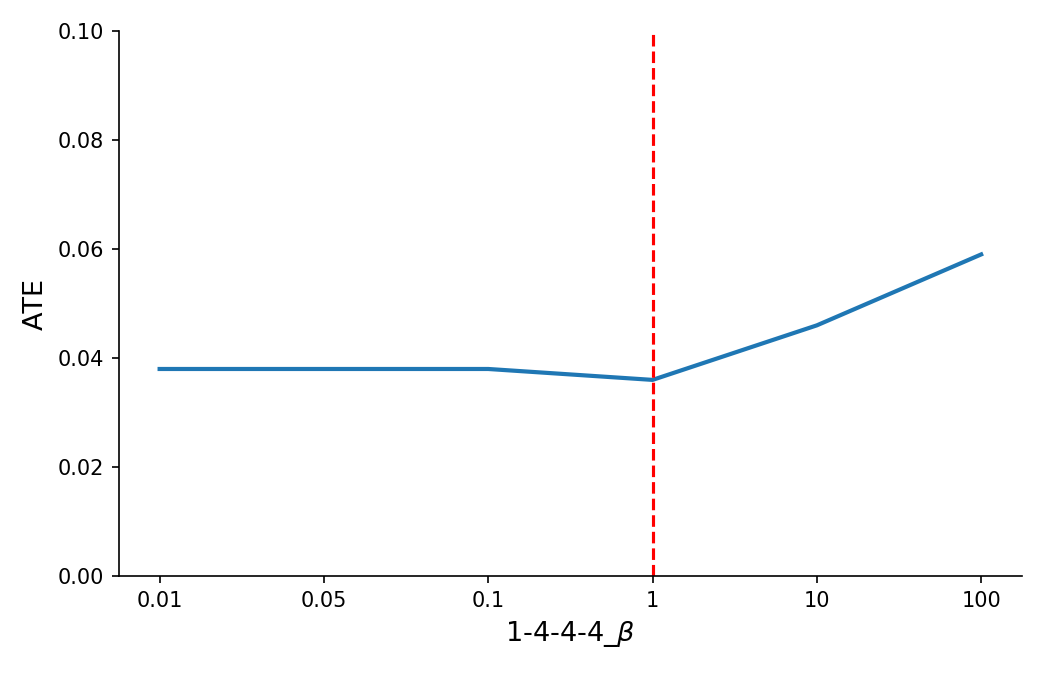}
\end{minipage}\hspace{0.05\textwidth}%
\begin{minipage}[b]{0.24\textwidth}
\centering
\includegraphics[width=\linewidth]{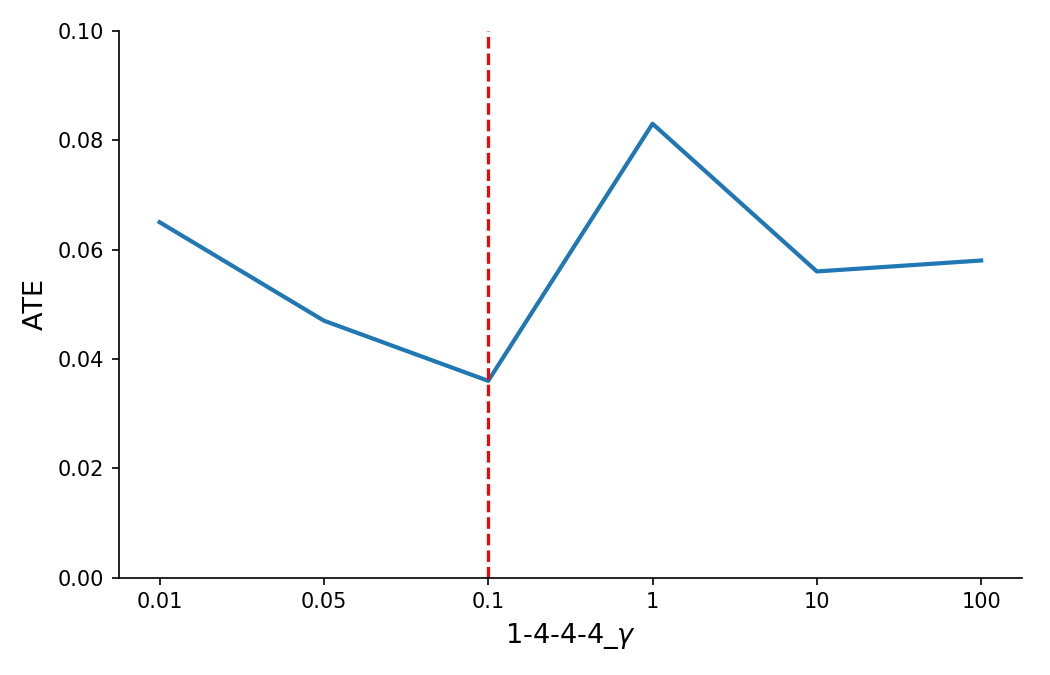}
\end{minipage}

\vspace{0.5em}
\begin{minipage}[b]{0.24\textwidth}
\centering
\includegraphics[width=\linewidth]{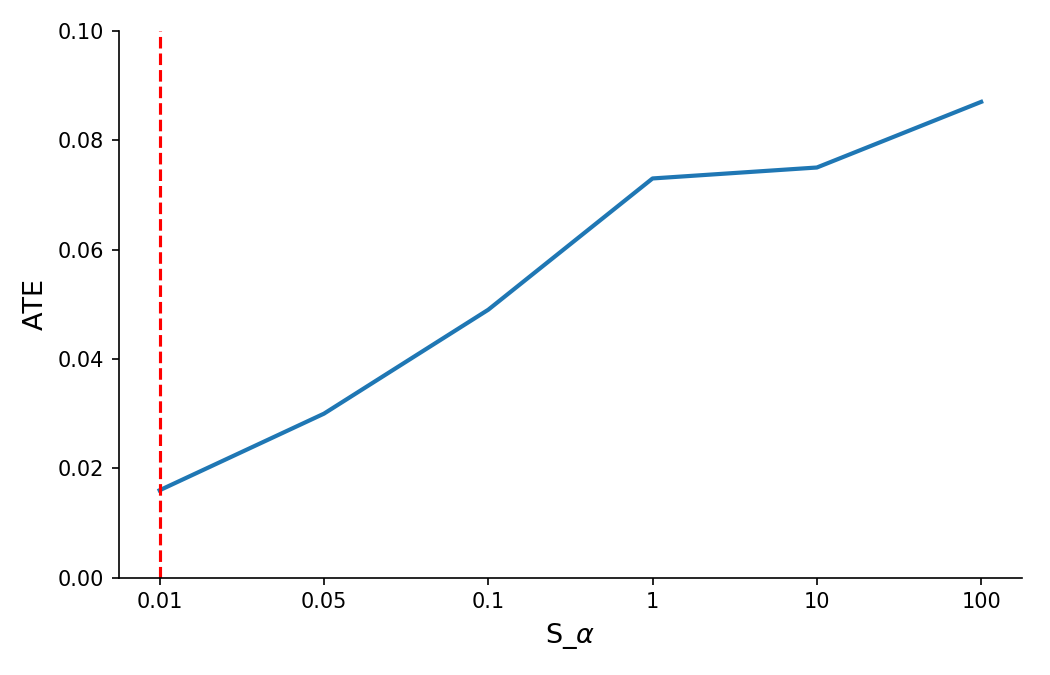}
\end{minipage}\hspace{0.05\textwidth}%
\begin{minipage}[b]{0.24\textwidth}
\centering
\includegraphics[width=\linewidth]{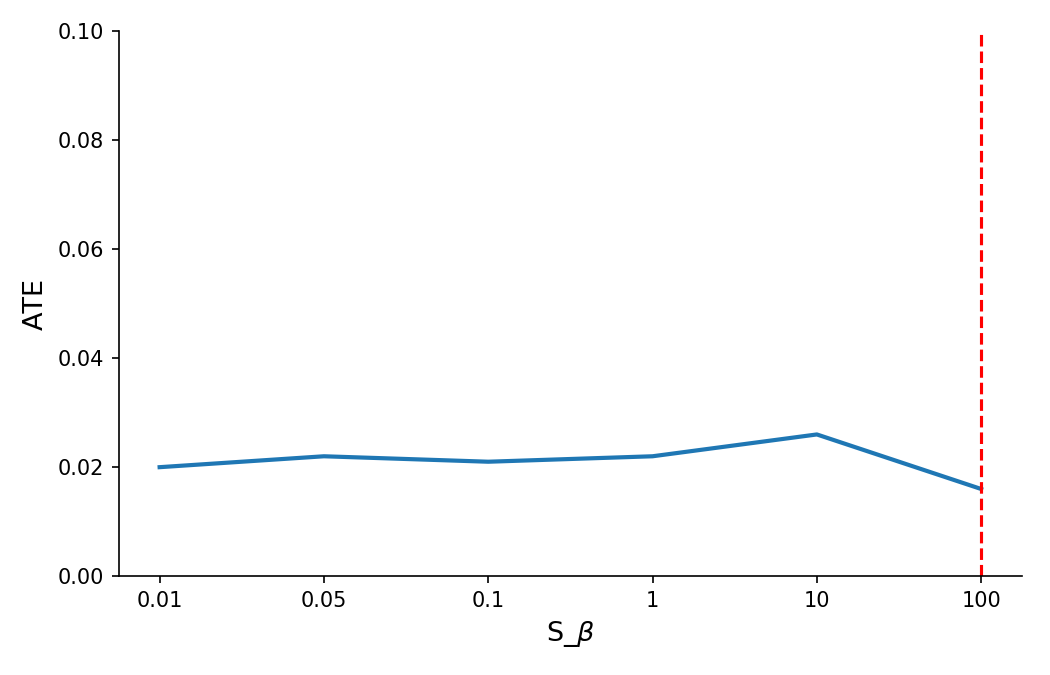}
\end{minipage}\hspace{0.05\textwidth}%
\begin{minipage}[b]{0.24\textwidth}
\centering
\includegraphics[width=\linewidth]{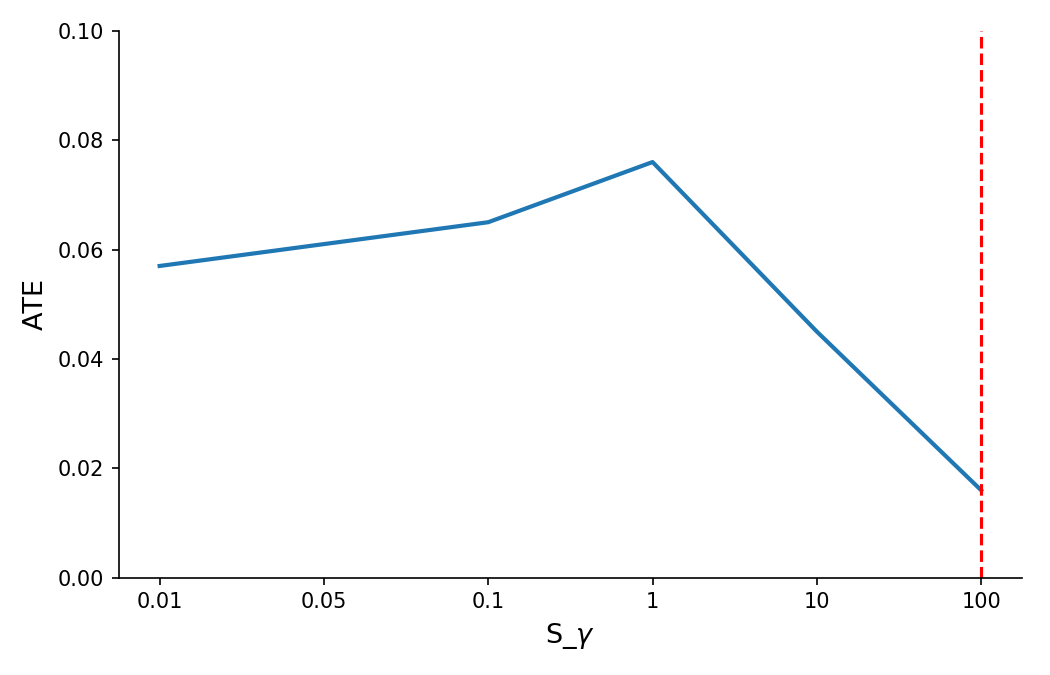}
\end{minipage}

\vspace{0.5em}
\begin{minipage}[b]{0.24\textwidth}
\centering
\includegraphics[width=\linewidth]{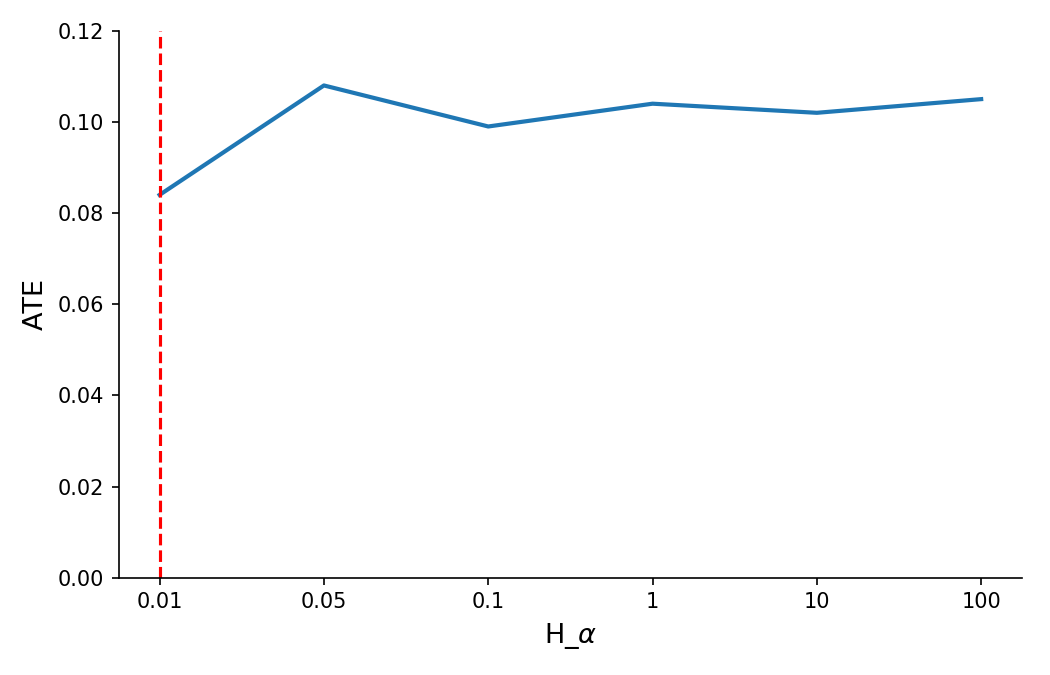}
\end{minipage}\hspace{0.05\textwidth}%
\begin{minipage}[b]{0.24\textwidth}
\centering
\includegraphics[width=\linewidth]{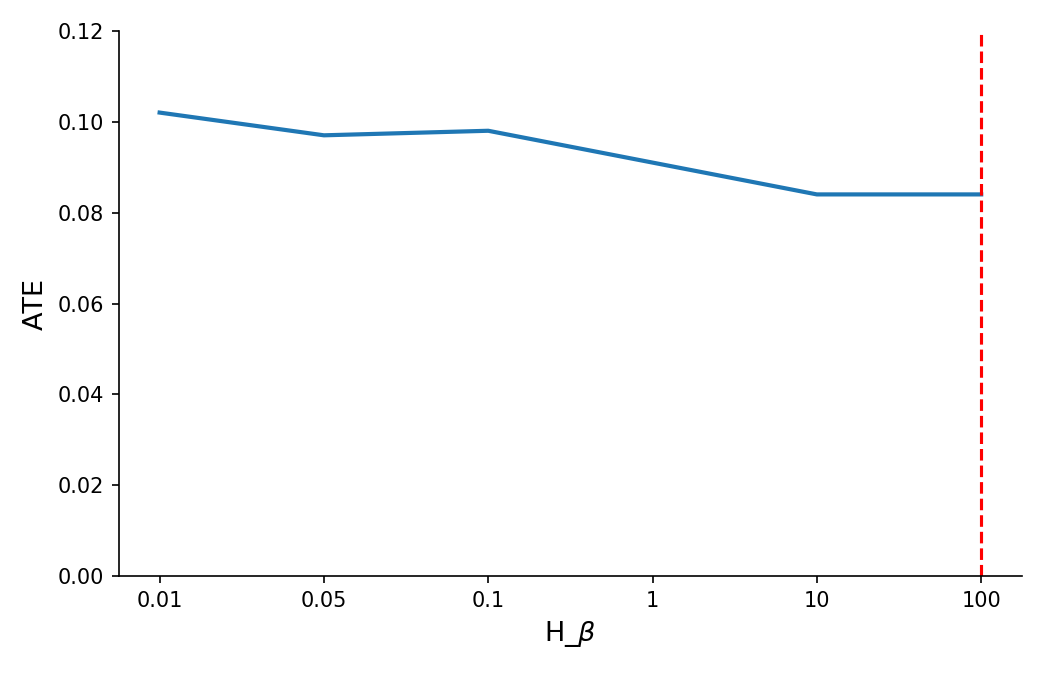}
\end{minipage}\hspace{0.05\textwidth}%
\begin{minipage}[b]{0.24\textwidth}
\centering
\includegraphics[width=\linewidth]{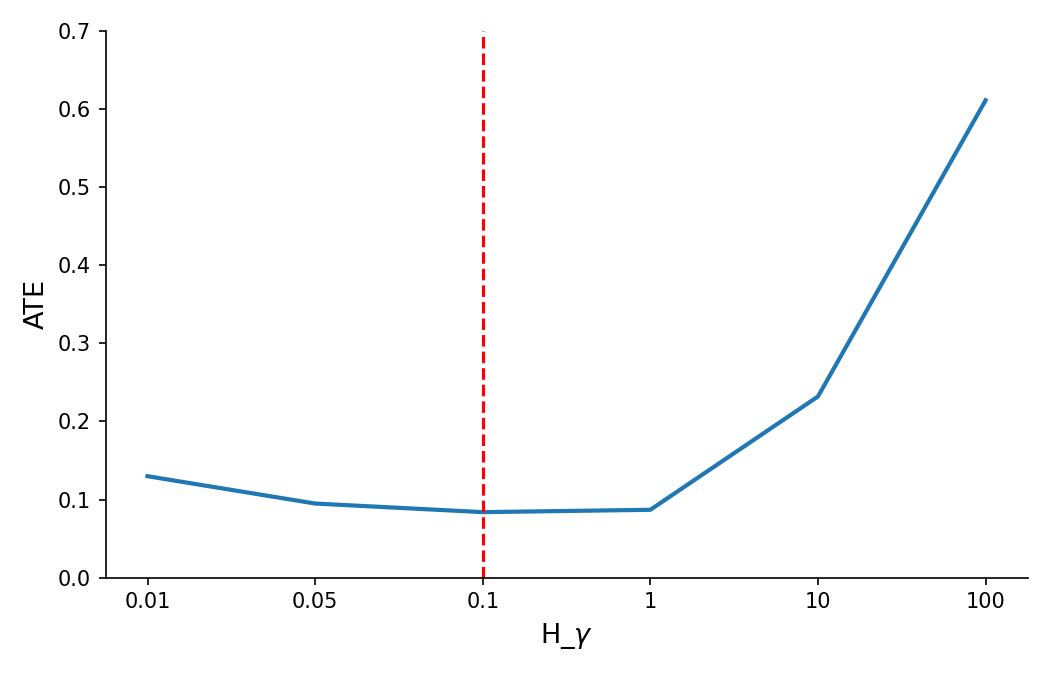}
\end{minipage}
\caption{Sensitivity of PCL-U to $\alpha$, $\beta$, and $\gamma$ on the synthetic 1-4-4-4, $s$,
and $h$ settings.}
\label{fig:sensitivity_appendix}
\end{figure}

Figure~\ref{fig:sensitivity_appendix} reports the sensitivity results for the
synthetic 1-4-4-4, $s$, and $h$ settings. There are noticeable that some hyperparameter settings are kind of big or low (e.g., $\alpha=0.05$ for 1-4-4-4, $\beta=100$ for $s$, and $\gamma=100$ for $s$) because the loss terms operate on different numerical scales. We therefore tune these weights to balance the contributions of the individual terms and ensure that each component receives meaningful gradient updates during training. Overall, PCL-U remains robust across a broad range of hyperparameter values.

\section{Implementation Details}
\label{sec:implementation_details}

All experiments in Python on a device equipped with an Intel Xeon w7-3455 CPU, an NVIDIA GeForce GT 710 GPU, and 256 GB of memory. We implement PCL-U in TensorFlow and jointly optimize all components using
Adam for 2,000 iterations with a mini-batch size of 128. The learning rate is
initialized at $1\times10^{-3}$ and multiplied by $0.97$ every 100 iterations.
The objective combines the MI-based proxy loss $\mathcal{L}_{\mathrm{proxy}}$,
the deep orthogonality regularizer $\mathcal{L}_{O}$, the outcome bridge loss
$\mathcal{L}_{h}$, and $\ell_2$ regularization on the bridge-network
parameters, with weights $\alpha$, $\beta$, $\gamma$, and $3\times10^{-7}$,
respectively. Each of the $Z$, $X$, and $W$ representation networks has three
hidden layers of width \texttt{dim\_in} and uses ELU activations. Its output
is normalized by dividing each sample's representation by its $\ell_2$
norm, with numerical stabilization (\texttt{normalization='divide'}).
The bridge network has three ReLU
hidden layers of width 80 and a linear output layer. We use zero-mean Gaussian
weight initialization with scale 0.1 and disable dropout by setting both keep
probabilities to 1. Table~\ref{tab:hyperparams} reports the dataset-specific
settings. Here, \texttt{scale} is the RBF kernel length scale,
\texttt{h\_split} indicates whether treatment-specific bridge networks are
used, \texttt{batch\_norm} indicates whether batch normalization is enabled,
and \texttt{outcome} specifies whether the outcome is continuous or binary.

For fixed network sizes and proxy dimensions, a mini-batch
optimization update with $b$ samples requires $O(b^2)$ time
and auxiliary memory due to the pairwise kernel computation
in $\mathcal{L}_h$. Compared with NMMR using prespecified
proxies, PCL-U incurs additional computation and parameter
storage for the proxy encoders and variational networks.
Full-loss evaluation on $m$ samples similarly requires
$O(m^2)$ pairwise computation and auxiliary memory.

\begin{table}[H]
\centering
\caption{Hyperparameters.}
\label{tab:hyperparams}
\footnotesize
\setlength{\tabcolsep}{6pt}
\renewcommand{\arraystretch}{1.0}
\begin{tabular}{@{}ccccccc@{}}
\toprule
\textbf{Hyperparameters} & \textbf{IHDP} & \textbf{Jobs} & \textbf{Twins} & \textbf{1/2/8-4-4-4} & \textbf{s} & \textbf{h} \\
\midrule
\texttt{dim\_in} & $16$ & $64$ & $64$ & $16$ & $64$ & $64$ \\
$\alpha$ & $0.01$ & $0.01$ & $0.01$ & $0.05$ & $0.01$ & $0.01$ \\
$\beta$ & $10$ & $1$ & $100$ & $1$ & $100$ & $100$ \\
$\gamma$ & $10$ & $10$ & $1$ & $0.1$ & $100$ & $0.1$ \\
\texttt{scale} & $0.05$ & $0.05$ & $0.05$ & $0.05$ & $0.3$ & $0.3$ \\
\texttt{h\_split} & Yes & No & Yes & Yes & Yes & Yes \\
\texttt{batch\_norm} & No & Yes & Yes & No & Yes & Yes \\
\texttt{outcome} & continuous & binary & continuous & continuous & continuous & continuous \\
\bottomrule
\end{tabular}
\end{table}

\end{document}